# FearCaut-Qwen: Affective Steering in a Vision–Language Model Shifts the Decision Criterion for Hazard Assessment

Xiaoshan Zhou

School of Project Management, Faculty of Engineering, The University of Sydney, Sydney, NSW 2006, Australia, xiaoshan.zhou@sydney.edu.au

## Abstract

Vision–language models (VLMs) show great potential for damage assessment in the built environment after a disaster, but a recurring deficiency is that they are reluctant to declare a hazard, i.e., recall is low even when overall accuracy appears adequate. This study examines that deficiency by using signal detection theory (SDT) to decompose the decision behavior into perceptual capability and decision-criterion placement. We then propose a novel method for correcting the over-conservative decision policy, inspired by the finding that fear makes humans risk-averse, and ask whether an affective representation associated with fear can be causally manipulated to similarly alter a VLM's decision tendency. Using mechanistic interpretability, we localize a causally implicated affective circuit in the model and use activation steering to manipulate it while observing the effect on downstream prediction. The method is tested on a two-stage SeisMLLM pipeline built on Qwen2.5-VL-7B-Instruct, which flags only 27.0% of genuinely unsafe buildings on the published SeisMLLM-1K test split and never issues a false Red, an SDT criterion of $c = +1.354$, despite adequate evidence quality ($d' = 1.521$). An affective direction is localized on emotion-rich natural scenes containing no structural-damage leakage, causally validated by sparse-neuron knockout and distributed steering on held-out emotion data, and then injected into the building task. Fear-direction injection raises Red recall to 75.7% ($p < 0.001$) and subtracting the same direction suppresses Red predictions entirely, whereas norm-matched random ($p = 0.46$) and matched happiness ($p = 0.84$) directions show no significant effect. The mechanism is a shift in criterion ($\Delta c = -1.515$) while discrimination is not improved ($\Delta d' = -0.493$). Perception-stage intervention recovers 94.4% of the effect, and the sparse neurons that are causally necessary for the source task carry none of the transfer. These results show that VLM hazard decisions can be adjusted at inference time without retraining and demonstrate how mechanistic interpretability can be used to diagnose and control VLM decision behaviors in engineering applications.

**Keywords:** building damage assessment; vision–language model (VLM); decision criterion; signal detection theory; activation steering; mechanistic interpretability

# 1. Introduction

Vision–language models (VLMs) are rapidly becoming the engine of automated safety assessment in the built environment. Because a single model can read several photographs of the same asset, follow a natural-language instruction, consult retrieved precedent and regulatory text, and export a structured report together with a decision, VLMs are now deployed across tasks that previously required a separate model each. In post-earthquake inspection, systems such as SeisMLLM, SDA-Chat and MT-SDAChat convert multi-view building photographs into damage descriptions and safety postings graded against the ATC-20-1 rapid-evaluation standard [1–4]. This extends to construction, where VLMs recognize contextual hazards on site and relate detected conditions to regulatory requirements [5–7], to open-ended inspection questions about bridge components and damage [8], to triage of disaster imagery with report and resource recommendations [9], and to estimates of perceived safety from street-view imagery at city scale [10,11].

Despite the promise and the high accuracy reported in these studies, VLMs typically expose a problem that recognition accuracy does not describe: they are systematically reluctant to call a hazard. SeisMLLM, a post-earthquake building damage framework for example, reports 66.13% safety-evaluation accuracy, yet the reconstruction evaluated here issues an Unsafe/Red posting for only 27.0% of genuinely unsafe buildings while never issuing a false Red [2]. In a controlled comparison of multimodal large language models (LLMs) on construction hazard recognition, zero-shot recall was only 0.288 for Claude-3 Opus and 0.326 for GPT-o3, meaning that most real hazards went unflagged [7]. A commercially deployed vision-enabled safety agent left 28% of true hazards undetected on real site photographs [12]. Wildfire damage assessment with GPT-4o and Qwen likewise struggles to commit to intermediate and severe damage from single views [13]. The common signature is that a VLM demands a great deal of evidence before it will declare a condition unsafe. This is particularly problematic in an engineering context, where a missed hazard can be far costlier than an unnecessary inspection.

This study therefore aims to address that problem. We suspect that the challenge is not that the VLM is incapable of discriminating damaged from undamaged structures, but that its decision threshold is too conservative, and we therefore hypothesize that lowering the boundary it applies before issuing a hazard will address the problem. To reposition that threshold we propose an approach based on the observation that affect changes perceived risk and appraisal in human judgement [14–18], and we ask whether fear can similarly produce pessimistic risk estimates and risk-averse choice [19–21] in VLMs. To test this, we first localize affective representations inside a VLM, then steer the activation of the identified fear/threat circuit and observe whether the manipulation produces the expected behavioural change. To determine whether any behavioural change reflects a repositioned decision criterion, we analyse the decision behaviour using signal detection theory (SDT), which separates a shift of the safety decision criterion from a change in how well the VLM actually sees damage.

Mechanistic interpretability provides tools for locating and manipulating representations inside a network. The field established that computation in neural networks can be decomposed into features and the circuits connecting them [22,23], that transformer feed-forward layers act as key–value memories whose individual coordinates can be inspected [24], and that causal tracing can localise and edit the components supporting a specific behaviour [25]. More recently, sparse dictionary methods have shown that interpretable features can be extracted at production scale [26], and the same toolkit has been extended to multimodal models, where visual tokens can be decoded in vocabulary space and their contribution ablated [27]. Two empirical findings bear on the feasibility of the proposed method. First, high-level behaviour can be changed at inference by adding a direction to the residual stream, without touching a weight [28–30]. Second, affect in particular appears to be linearly represented and causally manipulable: sentiment occupies a single direction in activation space [31]; emotion-selective neurons exist and their removal selectively impairs emotion prediction [32,33]; probing plus activation patching localises emotion inference to identifiable components [34]; and complete emotion circuits can be isolated and modulated to control expressed emotion [35]. VLM-specific controllers have been built for instruction adherence, refusal and embodied action [36–39]. What has not been established, and is central to demonstrating the efficacy of our proposed method, is whether manipulating an emotion-relevant neural circuit can change the decision-making of an unrelated engineering hazard assessment task, and, if so, whether the behavioural change reflects criterion movement or improved discrimination.

We close that gap with FearCaut-Qwen, demonstrated on a post-earthquake building damage assessment task that reproduces SeisMLLM. The method localizes an affective direction in Qwen2.5-VL-7B on emotion-rich natural scenes that contain no structural damage, establishes its causal relevance on held-out emotion data by sparse-neuron knockout and distributed steering, freezes every selection decision, and then injects the frozen direction into the two-stage perception-and-reasoning assessment pipeline evaluated on the published SeisMLLM-1K test split. SDT decomposes every resulting behavioural change into discrimination, $d'$, and decision criterion, $c$, so that a gain in recall can be attributed to the mechanism that produced it.

The remainder of the paper is organised as follows. Section 2 reviews the affective modulation of safety judgement, and mechanistic interpretability and activation steering. Section 3 specifies the method. Section 4 reports the baseline criterion defect, source-domain causal validation, the frozen target intervention, the SDT decomposition, and the stage and scale dissociations. Section 5 discusses the findings against prior work and states the limitations. Section 6 concludes.

# 2. Literature review

## 2.1 The influences of affect on safety decision making

That affect influences human decision making has been established over several decades, through classical accounts including affect-as-information [40], the somatic marker hypothesis [41], the appraisal-tendency framework [19,20], risk-as-feelings [15], and the emotion-imbued choice model that integrates them [21]. In particular, Lerner and Keltner found that fear makes people more cautious and risk-averse, whereas anger, like happiness, increases perceived control and risk seeking [20,42]. The influencing affect may be integral, arising from the object being judged, or incidental, carried over from an unrelated source into the next judgement [43,44].

How fear in particular influences decisions has been studied closely, and researchers have found that fear produces more cautious decisions and demands more evidence before a judgement is committed [45,46], and that it can form a prior or action bias that transfers to subsequent decisions [42,47]. A meta-analysis of the influence of fear on risk taking found that fear was associated with decreased risky decision making and increased risk estimation [48]. Of most direct relevance to the present design, recent work separates the fear effect into two mechanisms, a change in the rate of evidence processing and a change in decision caution, by fitting drift-diffusion models to emotional decision tasks [45,46]; this decomposition is the human analogue of the discrimination-versus-criterion separation we apply to a VLM. The pattern is not universal, however, and a recent high-powered experiment reports that incidental emotion does not significantly shift risk preferences [49], which makes empirical validation of an emotion-based manipulation in a model necessary.

Neurophysiological work localizes the emotion-relevant circuitry and indicates the mechanism of influence. The amygdala mediates rapid threat appraisal and biases the interpretation of ambiguous stimuli toward threat; in a framing experiment, amygdala activity tracked susceptibility to the framing bias while orbital and medial prefrontal activity predicted resistance to it, which is a change in how identical evidence is converted into choice [50]. In financial risk taking, anticipatory nucleus accumbens activation preceded risk-seeking errors and anterior-insula activation preceded risk-avoidant errors, again a shift in decision tendency rather than in the information available [51]. Reviews of the modulatory circuits explicitly distinguish incidental affect that shifts the decision threshold from integral affect carried by the stimulus itself [52], and the anxiety literature shows that trait and state anxiety bias both attention toward threat and the interpretation of ambiguity, producing systematically more avoidant choice without any change in stimulus discriminability [53]. This evidence comes largely from psychological and gambling tasks, but in more naturalistic on-site hazard experiments researchers have also used wearable electroencephalography to investigate the neural activation and connectivity underlying hazard recognition [54,55], and have found that negative emotional states measurably change hazard-assessment performance [17,18]. Together these findings motivate asking whether an affect-relevant circuit also exists in a VLM and whether fear can likewise influence its decision behavior.

## 2.2 Mechanistic interpretability and activation steering in vision–language models

Mechanistic interpretability provides the lens through which we understand and manipulate the VLM's decision making, so the relevant terms are first clarified. A network's computation decomposes into features, directions in activation space that encode interpretable variables, and circuits, the subgraphs that compute them [22]. For transformers this was formalised through the residual stream: every block reads from and writes to a shared additive channel, so a representation can be characterised by the direction it occupies in that stream and interventions can be expressed as additions to it [23]. The feed-forward sublayers admit a complementary unit-level reading, in which the post-gating intermediate coordinates behave as key–value memories whose individual entries can be inspected and perturbed [24]. These are the two representational scales investigated in previous studies, a sparse set of feed-forward coordinates and a distributed direction in the residual stream, and both are localized and manipulated here.

The literature also shows that establishing that a representation exists is not the same as establishing that it matters. Linear probing asks whether a variable is decodable from activations at a given depth; it bounds where to search but is purely associational, because a probe can read information the model never uses. Causal methods close that gap. Causal tracing and localized editing identified the components supporting factual recall and demonstrated that editing them changes model behaviour [25], and ablation or knockout tests necessity by removing a candidate component and measuring the selective loss. Activation steering tests sufficiency by adding a direction derived from contrastive examples during the forward pass [28], and contrastive activation addition computes residual-stream directions from paired prompts and changes high-level behaviour while largely preserving general capability [29]. Because individual neurons are frequently polysemantic, sparse dictionary learning has been introduced to recover more monosemantic features at scale [26]. Extending the toolkit to VLMs raises the further question of where visual information resides: object-identification accuracy drops by more than 70% when object-specific visual tokens are ablated, visual token representations become progressively interpretable in vocabulary space across layers, and object information is extracted at the final token position in a manner parallel to text-only factual recall [27].

Control methods for VLMs have followed quickly. SteerVLM learned a modulation module amounting to 0.14% of the base model's parameters and applied token-specific, layer-adaptive steering [36]. AutoSteer selected a safety-relevant layer automatically using a safety-awareness score, trained an intermediate prober, and triggered a refusal head only for risky multimodal inputs [37]. Interpretability-guided steering has been applied to vision–language–action models to change embodied policy at inference [38], and head- and neuron-level causal attribution has been combined in multitask VLMs [39]. These studies guide the steering experiments in our analyses.

Whether an affective influence on decision making analogous to the human case exists inside an LLM or VLM is a question that has recently become answerable, and the evidence is now substantial. Broekens et al. [56] showed that fine-grained affective processing emerges from LLMs without task-specific training, including meaningful valence, arousal and dominance representations and appraisal-based emotion elicitation consistent with the OCC model. At the representational level, sentiment is encoded along a single linear direction in activation space that generalizes across tasks and is summarized at syntactically neutral positions [31], and style and emotion vectors computed from recorded activations can steer generation toward a target affect when added during the forward pass [30]. At the unit level, Lee et al. [32] ranked emotion-sensitive neuron groups across six basic emotions and showed that removing the selected units selectively impairs emotion prediction, and Zhao et al. [33] extended neuron-level selection, suppression and amplification to large audio–language models with causal validation. At the circuit level, Tak et al. [34] combined linear probing with activation patching to localize emotion inference to mid-layer attention components and steered latent appraisal dimensions such as self-agency with theoretically consistent results, and Wang et al. [35] isolated complete emotion circuits comprising neurons and attention heads, validating them by both ablation and enhancement and achieving near-perfect control of expressed emotion. This body of work establishes that affective representations in these models are real, localizable and causally manipulable. The remaining gap is whether such a representation can change a decision criterion on a subsequent engineering safety evaluation task, which is the transfer question we pose.

# 3. Methodology

## 3.1 Overview

The overview of the research methodology is shown in Figure 1. First, we examine whether affective representations, identified here as a neural circuit, can be localized and causally validated in a Qwen model. The term *circuit* is used to denote a set of causally implicated components identified at three scales: a contiguous band of decoder layers in which affect is linearly decodable; a sparse set of SwiGLU intermediate coordinates whose attenuation selectively impairs source-task emotion recognition; and a distributed residual-stream direction whose addition or subtraction changes behaviour. Based on the identified fear/threat circuit, we then manipulate it and investigate whether it can bidirectionally change building damage assessment decisions. If it can, we examine whether any such change results from a gain in discrimination capability or from a shift of the decision criterion.

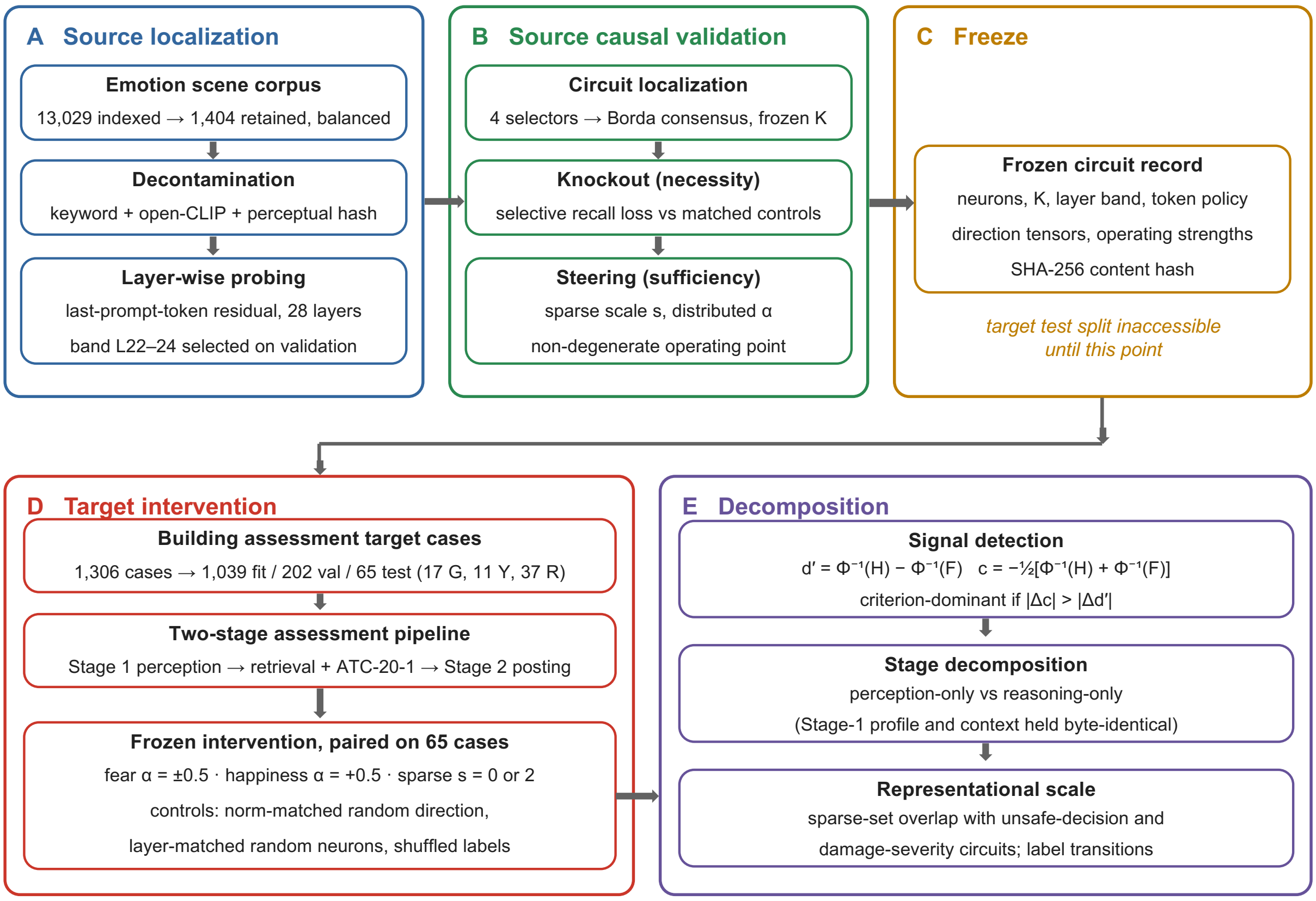

**Figure 1.** Overview of FearCaut-Qwen. Layer-wise probing localizes the candidate layer band; circuit localization ranks sparse neurons and estimates distributed directions; knockout and steering establish causal necessity and behavioural sufficiency on held-out emotion data; the resulting circuit is content-hashed and frozen; only then is the intervention transferred to the published SeisMLLM-1K test split and decomposed.

In our analysis, we separate three standards of evidence. Association means that affect is linearly decodable from activations. Sparse causal necessity means that selective attenuation impairs the target source class more than every matched control. Distributed behavioural sufficiency means that a non-degenerate direction changes source predictions in the intended direction. Details of the analyses are provided in the subsequent subsections.

## 3.2 Building damage assessment dataset

SeisMLLM-1K is a case-level, multi-view benchmark for post-earthquake building safety evaluation graded against ATC-20-1 [1,2]. Its public release contains 1,306 building cases and 3,421 images from 17 earthquakes of magnitude 6.1 to 8.0 recorded between 2010 and 2024. Cases span varied structural systems and geographic regions, and each pairs images with earthquake metadata, construction type, story count, six hazard conditions, per-image descriptions, estimated damage and a Green/Yellow/Red posting (Figure 2). The dataset contains 301 Green, 265 Yellow and 740 Red cases.

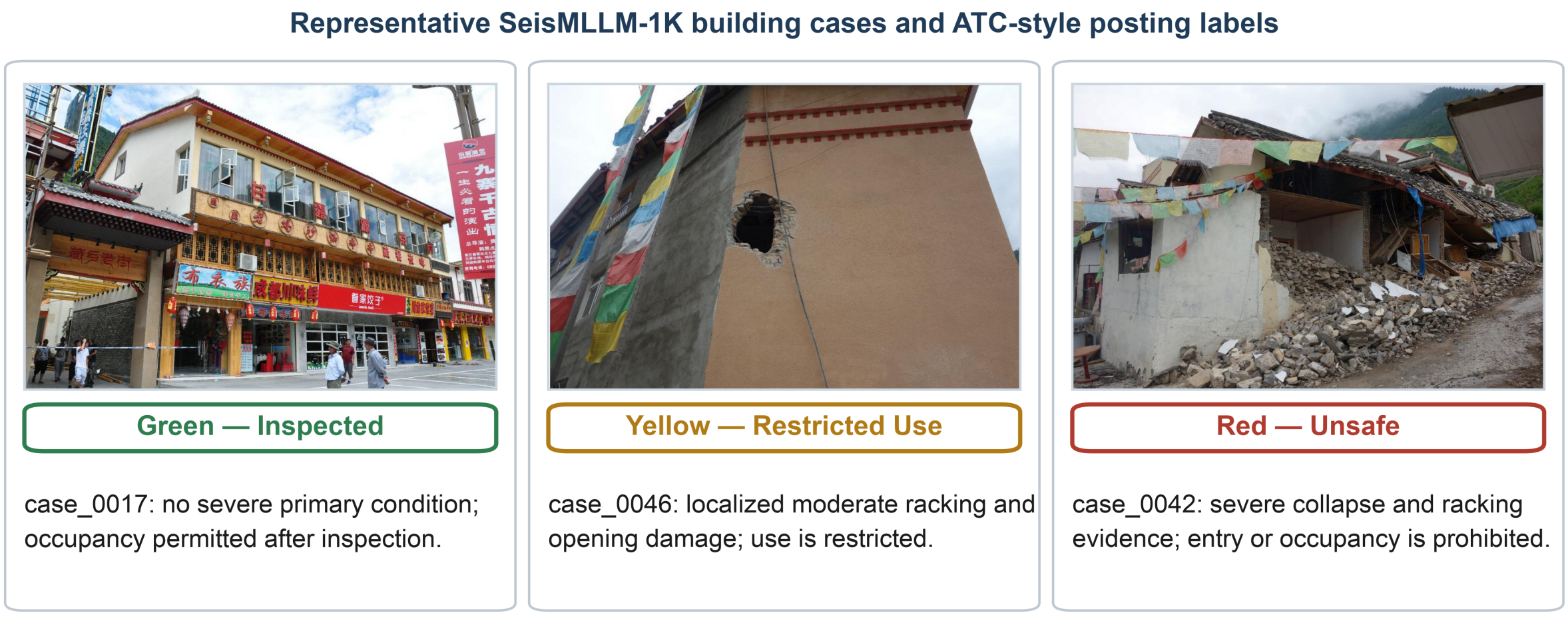

**Figure 2.** Representative SeisMLLM-1K cases. Green/Inspected, Yellow/Restricted Use and Red/Unsafe examples illustrate the difference between an apparently intact building, a localized falling or racking hazard, and severe collapse. Images and labels are from the public SeisMLLM-1K release [2].

The published split contains 1,241 training cases and 65 test cases. We retained the published test split unaltered. Fifteen percent of the training set was reserved for validation, stratified by earthquake event and safety label with duplicate-image blocks kept intact, yielding 1,039 train-fit, 202 validation and 65 test cases. The test split contains 17 Green, 11 Yellow and 37 Red cases.

### 3.3 Signal detection formalization in building damage assessment

SDT describes how an observer categorizes signal from noise under uncertain evidence [57,58]. The observer accumulates evidence in favour of the signal until a threshold is reached. The central contribution of SDT is that it separates the assessment decision into two components: sensitivity ($d'$), which describes how well the observer can distinguish signal from noise, and decision criterion ($c$), that is, how much evidence the observer requires before declaring that a signal is present. SDT has been used extensively to characterise human construction hazard recognition [55,59].

In the building damage assessment situation, we define Red/Unsafe as the signal class and Green or Yellow as the noise class. Accordingly, a hit is a Red prediction for a true Red case, and a false alarm is a Red prediction for a true Green or Yellow case. With $H$ the hit rate, $F$ the false-alarm rate and $\Phi^{-1}$ the standard-normal quantile function,

$$d' = \Phi^{-1}(H) - \Phi^{-1}(F) \quad (1)$$

$$c = -\frac{1}{2}[\Phi^{-1}(H) + \Phi^{-1}(F)] \quad (2)$$

Here $d'$ quantifies the separation of the internal evidence distributions for unsafe and non-unsafe buildings, and $c$ locates the response boundary relative to the midpoint of the two distributions (Figure 3). A lower $c$ ($c < 0$) denotes a more liberal criterion and a greater readiness to issue Red; a higher $c$ ($c > 0$) denotes an evidence-demanding and conservative criterion that is more permissive about occupancy in damaged buildings. To avoid the inverse-normal transformation becoming infinite at rates of 0 or 1, a symmetric log-linear (Hautus) correction is applied to both rates whenever either is degenerate [60]:

$$H = \frac{n_{\text{hit}} + 0.5}{n_{\text{signal}} + 1}, \quad F = \frac{n_{\text{fa}} + 0.5}{n_{\text{noise}} + 1} \quad (3)$$

The correction is applied to hits and false alarms symmetrically so that it cannot itself bias $c$. We also report the likelihood ratio at criterion, $\beta = exp\,[\frac{1}{2}(z_F^2 - z_H^2)]$, and the non-parametric sensitivity index $A'$, as robustness checks that do not assume equal-variance Gaussian evidence.

A paired comparison before and after intervention yields $\Delta d' = d'_{\text{int}} - d'_{\text{base}}$ and $\Delta c = c_{\text{int}} - c_{\text{base}}$. The dominant component is whichever of $|\Delta c|$ and $|\Delta d'|$ is larger, and we summarise the balance by the ratio

$$\rho_{c/d} = \frac{|\Delta c|}{|\Delta d'|} \quad (4)$$

A change is classified as a criterion shift without discrimination change when $|\Delta d'| \leq \tau$ and $|\Delta c| > \tau$, with the tolerance fixed at $\tau = 0.10$. Classifications in which $|\Delta d'| > \tau$ are reported with the dominant component stated explicitly, so that a large criterion movement accompanied by a smaller discrimination change is not reported as a discrimination result.

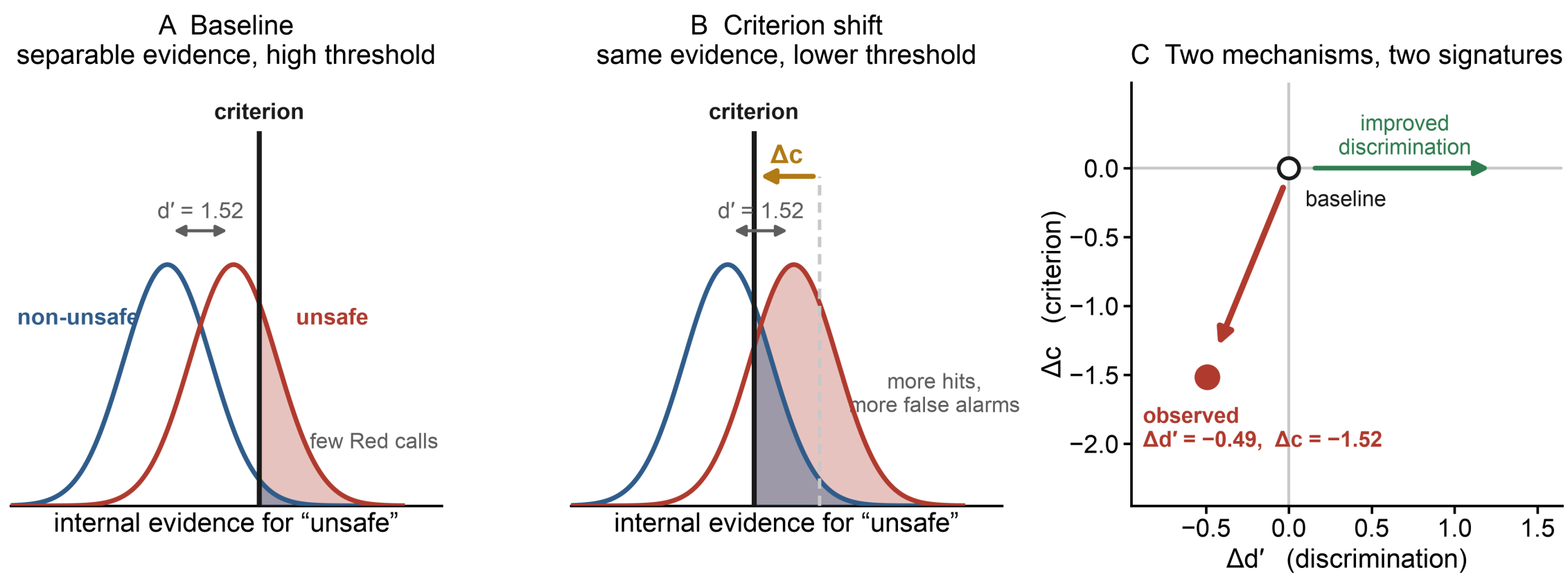

**Figure 3.** Discrimination and decision criterion in the Red-versus-non-Red building damage assessment decision. (A) The baseline: the evidence distributions for unsafe and non-unsafe buildings are separable, but the criterion sits far to the right, so few Red calls are made. (B) The same distributions with a lower criterion: hits and false alarms both increase, with no change in the underlying evidence. (C) The two mechanisms produce orthogonal signatures in the $\Delta d'$–$\Delta c$ plane; the observed effect of FearCaut-Qwen is marked.

The decomposition of sensitivity and decision criterion is important because, as shown in Figure 3, if a benchmark contains many Red cases, a liberal criterion can raise recall, accuracy and macro-F1 while discrimination is constant or worse. Classification metrics alone therefore cannot distinguish a model that has really learned something from a model that has simply become readier to flag everything Red. In this study we accordingly report conventional classification, ordinal error, calibration, $d'$ and $c$ together, to better understand the VLM's decision-making behaviour.

## 3.4 VLM model and the two-stage assessment pipeline

We adopted Qwen/Qwen2.5-VL-7B-Instruct as the frozen VLM backbone [61]. Qwen2.5-VL couples a vision transformer and a visual merger to a decoder-only language model, uses dynamic image resolution and multimodal rotary position encoding to preserve spatial structure, and supports object grounding and structured output. These properties matter for building assessment because one case may contain a variable number of exterior, interior and detail views, and the output must follow a fixed inspection schema. The instantiated language decoder has $L = 28$ blocks, hidden width $d = 3{,}584$ and SwiGLU intermediate width $d_{\text{ff}} = 18{,}944$. No parameter is fine-tuned: FearCaut-Qwen intervenes only on forward-pass activations, and all primary runs use deterministic greedy decoding.

Experiments ran on an Apple M5 Pro with 48 GB unified memory under macOS 26.6.2, with Python 3.12.14, PyTorch 2.14.0, Transformers 5.16.1 and scikit-learn 1.9.0. Computation used the Metal Performance Shaders backend, bfloat16 and batch size 1.

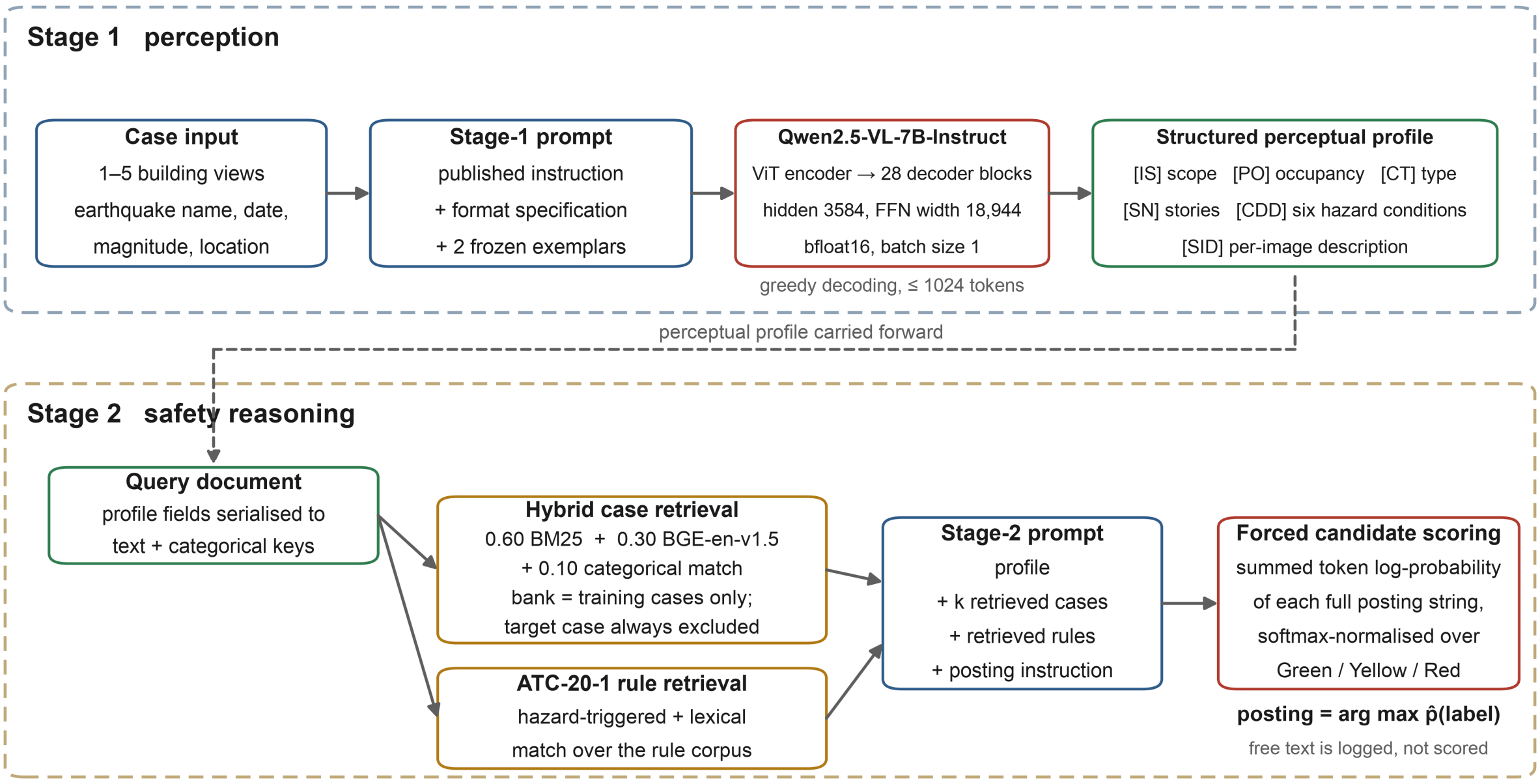

**Figure 4.** The reconstructed two-stage SeisMLLM pipeline. Stage 1 turns multi-view photographs and earthquake metadata into a structured perceptual profile. Stage 2 serialises that profile into a query document, retrieves verified cases by a hybrid lexical–semantic–categorical score and ATC-20-1 rules by hazard trigger and lexical match, assembles the reasoning prompt, and scores the three complete posting strings by summed token log-probability. The posting is the arg max of the normalised candidate probabilities; free-text generation is logged but is not the primary decision.

We reproduced SeisMLLM using its two-stage pipeline (Figure 4). In Stage 1 (perception), all images belonging to a case are placed in a single prompt under a 4,096 visual-token budget, with 64 to 1,280 tokens allocated per image according to its resolution. The largest case, with 36 images, consumed 4,016 prompt tokens at 88,592 pixels per image. We used the published instruction from the release records, with the literal `<image>` placeholders stripped because the processor inserts vision placeholders itself from the message content, thus retaining them would double-count the visual positions. To the published instruction we append an explicit output schema and two fixed worked examples, because zero-shot prompting did not reliably elicit the required structure. The two exemplars are drawn from the train-fit split only, are restricted to single-image cases with different postings so that the format is demonstrated without biasing the label prior, and are selected once under seed 37 and then held constant for every run in the study.

The model returns a structured profile with six fields: image scope [IS], primary occupation [PO], construction type [CT], stories over ground [SN], condition of damage [CDD] and per-image descriptions [SID]. The [CDD] field enumerates six hazard conditions in fixed order: collapse or off-foundation, building or story leaning, racking or other structural damage, chimney/parapet/falling hazard, ground movement, and other hazards, each with a severity drawn from {not visible, minor or none, moderate, severe} followed by parenthesised visual evidence. Stage 1 allows 1,024 new tokens.

During Stage 2, the perceptual profile is converted into a posting using two retrieval channels and an ATC-20-1-grounded instruction. The parsed profile is first serialised into a query document $q$ containing the textual fields and the categorical keys. Case retrieval over the train-fit bank $\mathcal{B}$ combines three components. The lexical component is a BM25 score, the semantic component is the inner product of BGE-en-v1.5 sentence embeddings, and both are min–max normalised across the candidate bank so that they occupy a common scale before weighting:

$$\tilde{s}(q,b) = \frac{s(q,b) - \min_{b'\in\mathcal{B}} s(q,b')}{\max_{b'\in\mathcal{B}} s(q,b') - \min_{b'\in\mathcal{B}} s(q,b')} \quad (5)$$

The categorical component is a weighted field-agreement score. Three of the four fields are matched exactly, while the story count is matched on a graded scale because a one-storey discrepancy between a three-storey and a four-storey building is not the same kind of mismatch as a wood frame reported as unreinforced masonry:

$$g_{\text{cat}}(q,b) = \frac{1}{\sum_k w_k}\left[\sum_{k\in\{\text{CT,IS,PO}\}} w_k \mathbb{1}\left[q_k = b_k\right] + w_{\text{SN}} \max\left(0, 1 - \frac{|q_{\text{SN}} - b_{\text{SN}}|}{3}\right)\right] \quad (6)$$

with field weights $w_{\text{CT}} = 0.60$, $w_{\text{IS}} = 0.30$, $w_{\text{SN}} = 0.05$ and $w_{\text{PO}} = 0.05$. The three channels are then combined with the published hybrid weights:

$$S(q,b) = 0.60\, \tilde{s}_{\text{BM25}}(q,b) + 0.30\, \tilde{s}_{\text{BGE}}(q,b) + 0.10\, g_{\text{cat}}(q,b) \quad (7)$$

The three highest-scoring cases are retrieved. The query case is always excluded from its own retrieval, and the bank contains train-fit cases only, so no test-case ground truth can enter the Stage-2 context through retrieval. Rule retrieval operates over a 23-rule reconstruction of ATC-20-1 evaluation guidance, selecting rules by hazard trigger, for example, a severe racking report activates the racking rules, and by lexical match against the query document.

A deterministic reference rule is also computed for diagnostic purposes: Red if any primary condition (collapse, leaning or racking) is severe; otherwise Yellow if a primary condition is moderate or a secondary condition is severe; otherwise Green. This rule agrees with the published labels for 86.9% of train-fit, 85.1% of validation and 81.5% of test cases. It supports retrieval and circuit analysis but is not reported as the model's prediction.

## 3.5 Forced candidate scoring and the decision rule

Free-text parsing is an unreliable decision rule under intervention, because a steered model may change its phrasing without changing its judgement, or fail to emit a parseable posting at all. The primary decision rule is therefore a forced-choice score over the three complete posting strings $\mathcal{Y} =$

{“(1) Inspected (Green)”, “(2) Restricted Use (Yellow)”, “(3) Unsafe (Red)”}. For a Stage-2 prompt with token sequence $x_{1:n}$ and an assistant prefix fixing the next tokens to be the posting itself, each candidate $y$ is tokenised to $y_{1:m_y}$ and scored by its summed conditional log-probability:

$$s_y = \sum_{j=1}^{m_y} log p_\theta \left(y_j \mid x_{1:n}, y_{1:j-1}\right) \quad (8)$$

Because the candidates are multi-token and of unequal length, Eq. (8) is a sequence log-likelihood rather than a single-token logit comparison. Normalised candidate probabilities follow from a softmax over the three scores,

$$\hat{p}(y) = \frac{exp(s_y)}{\sum_{y' \in \mathcal{Y}} exp(s_{y'})} \quad (9)$$

and the posting is $\hat{y}$ =\$arg$max$_y \hat{p}(y)$. These normalised probabilities also supply the scores used for the Brier score, expected calibration error and one-versus-rest AUC. Free-text Stage-2 generation is still produced and logged under a 256-token budget, and its parsed label is recorded for comparison. This matters for interpretability as well as robustness: because $\hat{p}$ is defined on identical candidate strings in every condition, we can distinguish a change in $\hat{p}$ attributable to the intervention from a change in surface wording.

We deliberately keep the two-stage design because it supports a controlled decomposition that a single-pass model cannot. Perception-only steering changes the evidence report before unmodified retrieval and reasoning. Reasoning-only steering reuses the baseline Stage-1 text, the parsed profile, the retrieved cases, the retrieved rules and the complete Stage-2 context byte for byte, so that any change must have occurred after the evidence and the retrieved context were fixed.

## 3.6 Source emotion dataset and decontamination

FindingEmo was used to elicit affective responses in the model backbone. The selection of this dataset had to satisfy two requirements. It must evoke a usable affective representation in the backbone, and it must share no structural-damage content with the target task, because a circuit that is partly a rubble detector would transfer for entirely uninteresting reasons. FindingEmo meets the first requirement by construction, and, compared with most widely used emotion datasets (for example AffectNet, RAF-DB and, in the visual-emotion setting, EmoSet) that are built around facial expressions or single-object imagery [62–64], FindingEmo was constructed for whole-scene emotion recognition [65]. We selected it because the source-to-target domain shift is smaller when whole scenes rather than cropped faces are presented to a model whose target task is also whole-scene perception. To meet the second requirement, we implemented an explicit decontamination procedure to prevent target-data leakage.

FindingEmo's public release provides annotations for 25,869 images and a 1,525-image private test set. Candidate scenes were gathered by Internet image search over combinations of emotion, social-group and setting terms, with 1,041,105 collected images filtered before annotation, and 655 Prolific participants annotated the emotional gist of the whole scene. Labels comprise 24 Plutchik leaves grouped into eight families, valence on a −3 to +3 scale and arousal on a 0 to 6 scale. FindingEmo distributes annotations and source URLs instead of the image files themselves.

Because the 24 Plutchik leaves are too fine-grained for a causal design, we mapped them to four classes: happiness from Serenity, Joy and Ecstasy; sadness from Pensiveness, Sadness and Grief; fear/threat from Apprehension, Fear and Terror; and neutral from annotations with valence exactly 0 and arousal at most 2 (Figure 5). Neutral supplies the reference against which every contrast is computed; happiness tests positive valence; sadness was intended to test negative valence at lower activation; and fear/threat tests negative valence at high activation with an uncertainty and low-control appraisal [66–68]. These are class tendencies in the annotation distribution, not constraints imposed on individual scenes, and Section 3.6.1 reports the extent to which the intended valence–arousal separation was actually realised. From 13,029 indexed annotations in our experimental retrieval, 3,600 candidates were sampled at 900 per class under seed 37, of which 2,457 images were successfully downloaded.

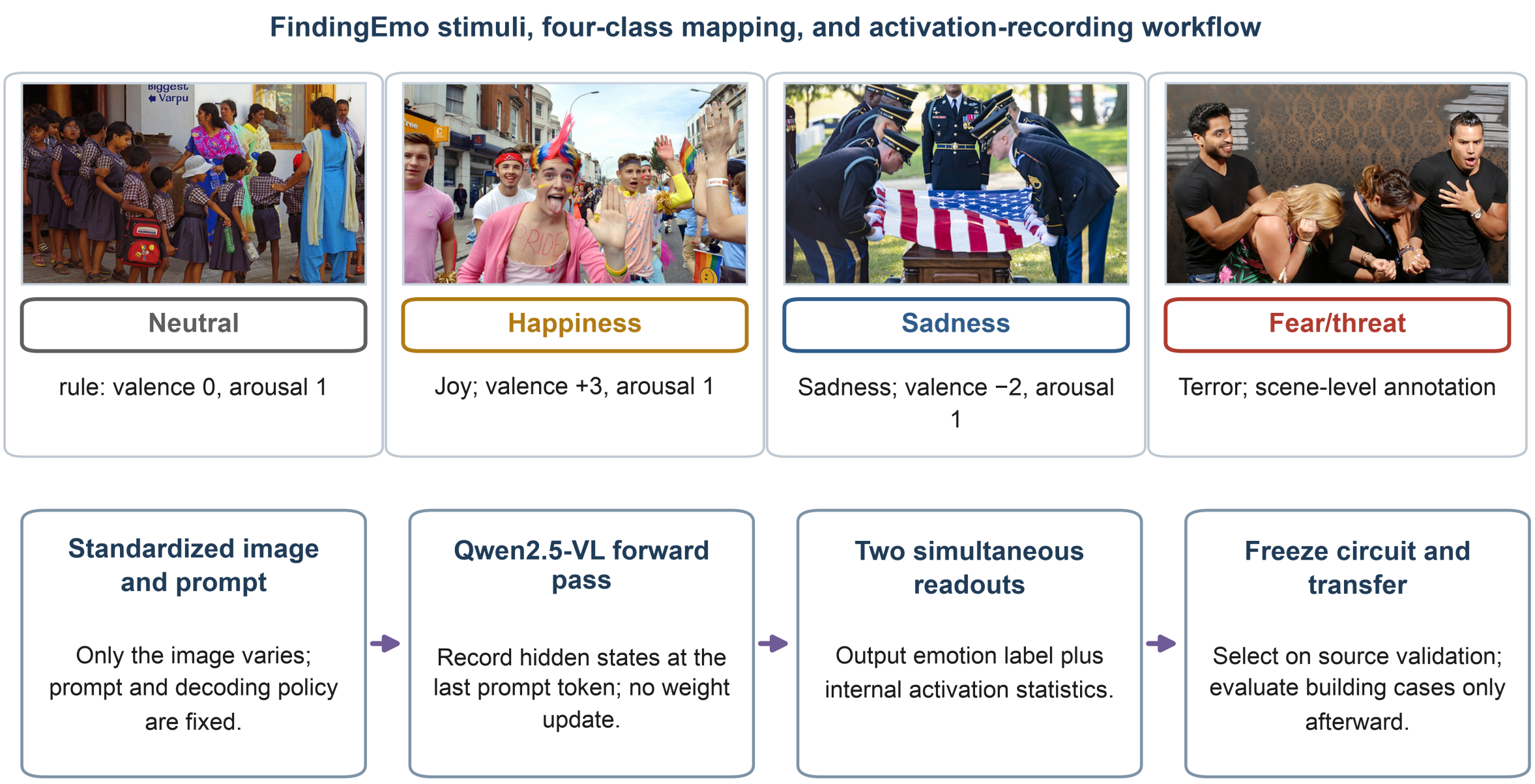

**Figure 5.** FindingEmo examples and the source-task workflow. URL-indexed scenes illustrate the four mapped source classes. A fixed four-choice prompt is paired with each image; one forward pass yields both an output label and internal activation measurements; the selected circuit is frozen before any target evaluation. The presentation of images to the VLM is analogous to the presentation of controlled stimuli in a human experiment [65].

Decontamination comprised three independent screens, summarised in Figure 6(A). First, metadata, scrape query, tags and labels were checked against a 61-term vocabulary covering earthquakes, damaged or collapsed buildings, rubble, disaster response, construction sites, inspection, war, wildfire and flood. Second, open-CLIP ViT-B-32 trained on LAION-2B scored each image against eight forbidden concepts and eight contrast concepts using contrastive image–text similarity [69,70]; any image whose normalised forbidden probability reached $\tau = 0.25$ was excluded. The Qwen model was deliberately not used as the filter, which would have made the decontamination circular. Third, perceptual hashing removed near-duplicates within FindingEmo and against every SeisMLLM image. Of the 2,457 downloaded images, 20 were removed as perceptual duplicates and 671 failed the semantic screen, leaving 1,766. No source image matched any target image. Exclusion was strongly class-dependent: 45% for fear/threat and 41% for sadness, against 7% for happiness and 17% for neutral. This is because negatively valenced web imagery contains disproportionately more disaster and conflict content.

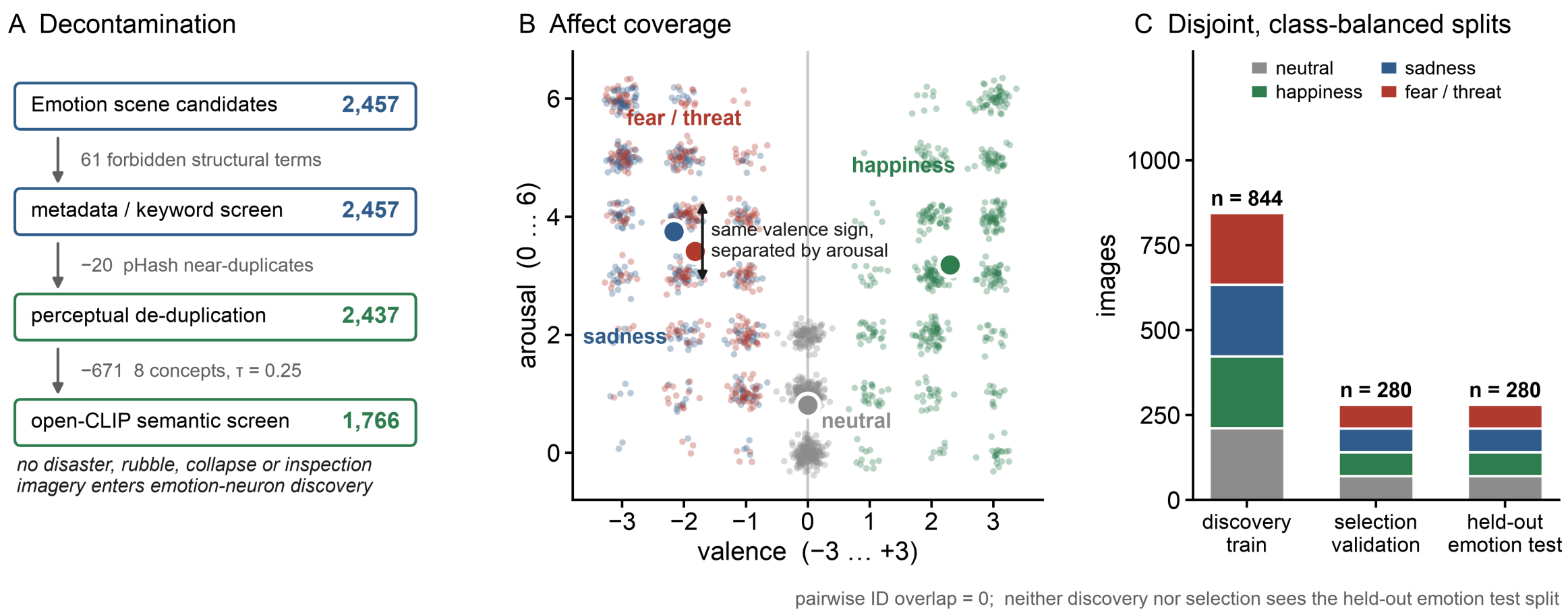

**Figure 6.** Construction of the emotion corpus. (A) The decontamination cascade, from 2,457 downloaded candidates to 1,766 retained images. (B) Valence and arousal coverage of the four retained classes; small points are individual annotated images (jittered for visibility, since both scales are integer-valued) and large points are the class centroids computed from those annotations. (C) The three class-balanced, pairwise disjoint splits used for discovery, selection and held-out confirmation.

### 3.6.1 Realised valence–arousal separation of the source classes

Because the intended fear-versus-sadness contrast depends on the two classes differing in arousal, we verified the realised separation in the retained annotations. The class centroids in Figure 6(B) are computed directly from the FindingEmo labels of the 1,766 retained images. Neutral sits at valence 0.00 and arousal 0.81 by construction, and happiness is clearly separated on valence (+2.30, arousal 3.18). The two negative classes, however, are separated mainly on valence, and in the opposite direction to the design intention: sadness has mean valence −2.16 and arousal 3.75, whereas fear/threat has mean valence −1.82 and arousal 3.41. Fear/threat is slightly less aroused than sadness in this corpus (Welch $t = -2.74$, $p = 0.006$, Cohen's $d = -0.20$) and slightly less negative in valence ($t = 5.95$, $p < 10^{-8}$, $d = 0.44$). The planned two-dimensional dissociation of negative valence from high arousal is thus not realised by these annotations. Any difference we observe later between fear/threat and sadness cannot thus be attributed to an arousal difference in the stimulus set, and the limitation compounds the separate failure of sadness to admit a usable steering strength reported in Section 4.3.

The retained images were balanced to 1,404 and partitioned into three pairwise disjoint, class-balanced splits (Figure 6C): 211 images per class for discovery (844), 70 per class for selection validation (280) and 70 per class for held-out emotion test (280). Pairwise identifier overlap was verified to be exactly zero. The discovery split is the only data from which neuron rankings and steering directions are computed; the selection-validation split is the only data on which $K$ and steering strength are chosen; the emotion-test split is evaluated once, as confirmation. Each retained image is paired with the same four-choice emotion prompt and decoding policy, so that the model serves simultaneously as classifier and as object of measurement.

## 3.7 Activation capture: neuron definition, token policy and streaming statistics

A sparse neuron is defined as one coordinate of the post-gating SwiGLU intermediate representation immediately before the MLP down-projection. For decoder block $\ell$ with input $x$,

$$a^{(\ell)} = \text{SiLU}\,(W_{\text{gate}}^{(\ell)}x) \odot (W_{\text{up}}^{(\ell)}x), \quad \text{MLP}^{(\ell)}(x) = W_{\text{down}}^{(\ell)}\, a^{(\ell)} \quad (10)$$

so that $a^{(\ell)} \in \mathbb{R}^{18{,}944}$ is exactly the vector the down-projection reads [24]. Module paths were discovered dynamically from the instantiated model rather than hard-coded, and a forward pre-hook on $W_{\text{down}}^{(\ell)}$ captures $a^{(\ell)}$ without altering it.

Pooling over token positions requires a single frozen policy, because mixing policies within one ranking would confound position with representation. Four policies were implemented, and the primary analyses use the last prompt token, which is identical in form across all source stimuli and precedes any emitted class token; this position is also where object-level visual information has been shown to be aggregated in VLMs [27]:

$$\bar{a}^{(\ell)} = \frac{1}{|\mathcal{T}|}\sum_{t\in\mathcal{T}} a_t^{(\ell)} \quad (11)$$

with $\mathcal{T}$ the selected positions. The first-answer-token policy was retained as a diagnostic, and it demonstrates why the choice matters: under that policy, balanced accuracy at layer 0 was already 0.661, because the observed token encodes the answer the model is about to give, whereas under the last-prompt-token policy layer 0 is at chance (0.250). Probing at the first answer token therefore reads the model's own output instead of its internal representation, and would have produced a spurious appearance of early-layer affect coding.

Full activation tensors are never retained for the whole corpus. Instead, per-neuron sufficient statistics are streamed on CPU in float64 using Welford updates, which are numerically stable in one pass:

$$n \leftarrow n+1, \quad \mu \leftarrow \mu + \frac{a-\mu}{n}, \quad M_2 \leftarrow M_2 + (a-\mu_{\text{old}})(a-\mu) \quad (12)$$

with $\sigma^2 = M_2/(n-1)$. The same recursion is maintained separately for each class. Three further statistics are accumulated: mean absolute activation, used later to build magnitude-matched controls; top-activation frequency, defined per layer as the proportion of samples in which a neuron falls in the top 1% of that layer's coordinates; and a 10-bin class-conditioned response histogram, binned on global quantile edges obtained from a warm-up pass. Raw per-sample activations are stored for the three-layer candidate band and for a 24-sample diagnostic subset used to validate the streaming statistics against exact computation.

## 3.8 Layer-wise affect probing and selection of the candidate band

Linear probing answers where affect information is available, which bounds where it is worth searching for causal structure. At each decoder-layer boundary, standardised residual vectors from the discovery split are entered into a class-balanced multinomial logistic regression. Standardisation uses discovery-split means and standard deviations applied unchanged to the other splits, so no target-split statistics leak into the transform. The inverse regularisation strength is selected from $C \in \{0.001, 0.01, 0.1, 1, 10\}$ on the selection-validation split by balanced accuracy, and the held-out emotion-test split is evaluated exactly once. Ridge regressions with $\alpha \in \{0.1, 1, 10, 100, 1000\}$ separately predict FindingEmo valence and arousal, scored by $R^2$. Confidence intervals for held-out performance use 2,000 bootstrap resamples of the test split.

The candidate layer band is then derived from validation scores only, by taking the longest contiguous run among the eight highest-scoring layers, with a minimum run length of three. This selected layers 22–24. The band is used in two ways later: it bounds the universe for the structural circuit comparison in Section 3.13, and it is the set of layers at which steering directions are defined.

A probe is an association test. Successful decoding shows that information is linearly available at a layer; it does not show that the layer is necessary, and it says nothing about any individual coordinate [22,23]. FearCaut-Qwen therefore proceeds to intervention before transferring anything.

### 3.9 Sparse neuron selectors and Borda consensus

We used four complementary selectors to rank neurons for each target emotion $e$ against the neutral reference, each computed on the discovery split alone. Each selector captures a different sense in which a neuron may be emotion-sensitive, and their disagreement is informative rather than a defect.

The contrast-effect selector is a standardised mean difference between the target class and the neutral reference, with a pooled standard deviation, that is, Cohen's $d$:

$$\mathrm{CE}_e(\ell, i) = \frac{\bar{a}_e(\ell, i) - \bar{a}_{\mathrm{neu}}(\ell, i)}{s_{\mathrm{pool}}(\ell, i)}, \quad s_{\mathrm{pool}} = \sqrt{\frac{(n_e - 1)\sigma_e^2 + (n_{\mathrm{neu}} - 1)\sigma_{\mathrm{neu}}^2}{n_e + n_{\mathrm{neu}} - 2}} \quad (13)$$

The mean-deviation selector measures how far the class-conditional mean departs from the global mean in units of the global standard deviation, and so rewards neurons whose target response is unusual with respect to the whole corpus rather than only with respect to neutral:

$$\mathrm{MD}_e(\ell, i) = \frac{\bar{a}_e(\ell, i) - \bar{a}(\ell, i)}{\sigma(\ell, i)} \quad (14)$$

The frequency selector is insensitive to activation scale and asks instead how often a neuron is among the most strongly driven coordinates of its layer. With $f_c(\ell, i)$ the proportion of class-$c$ samples for which neuron $(\ell, i)$ lies in the top 1% of layer $\ell$, the score is the excess over the mean of the control classes, not over the global mean:

$$\mathrm{FR}_e(\ell, i) = f_e(\ell, i) - \frac{1}{C-1} \textstyle\sum_{c \neq e} f_c(\ell, i) \quad (15)$$

The entropy selector measures class selectivity of the whole response distribution rather than of its mean. From the class-conditioned histograms, let $P(b \mid c)$ be the normalised response distribution of neuron $(\ell, i)$ for class $c$ over $B$ bins, let $P(c \mid b) \propto P(b \mid c)$ be the implied class posterior in bin $b$, and let $\pi(b)$ be the bin mass averaged over classes and renormalised. The score is one minus the mass-weighted posterior entropy, normalised by its maximum and signed toward the target class:

$$\mathrm{EN}_e(\ell, i) = [1 - \frac{\sum_b \pi(b)\, \mathcal{H}\,(P(\cdot \mid b))}{logC}] \cdot \mathrm{sgn}\,(\textstyle\sum_b \pi(b) P(e \mid b) - \frac{1}{C}) \quad (16)$$

A neuron scores highly only if its response pattern is both informative about class identity and informative in favour of the target class. A Gaussian class-posterior approximation is used if histograms are unavailable.

Each selector produces a full ranking restricted to the candidate layers. Rankings are combined by Borda consensus over normalised ranks (Figure 7A). For selector $m$, let $r_m(\ell, i) \in [0,1]$ be the position of neuron $(\ell, i)$ in that selector's pool, normalised so that the best neuron scores 0 and the worst scores 1. The pool is deliberately much larger than $K$, at $\max(8K, 2048)$, so that consensus is computed over a broad candidate set instead of over four short lists. A neuron absent from a selector's pool is charged the worst possible normalised rank, so that selectors cannot be gamed by sparse agreement:

$$R_e(\ell,i) = \frac{1}{M}\left[\sum_{m:\,(\ell,i)\in\mathcal{P}_m} r_m(\ell,i) \;+\; (M - |\{m: (\ell,i)\in\mathcal{P}_m\}|)\right] \quad (17)$$

with $M = 4$. The $K$ neurons with the smallest $R_e$ form the sparse set $\mathcal{N}_e$. The size $K \in \{32, 64, 128, 256, 512\}$ is selected by balanced accuracy on the selection-validation split, using only the selected neurons as features, and is then frozen. This yielded $K = 256$ for fear/threat, $K = 128$ for happiness and $K = 256$ for sadness; the sweep and the frozen choices are shown in Figure 7B, and the layer distribution of the selected sets in Figure 7C. Stability is quantified two ways: 50 bootstrap resamples of the discovery split, scored by Jaccard overlap against the full-sample set, and five stratified folds compared pairwise.

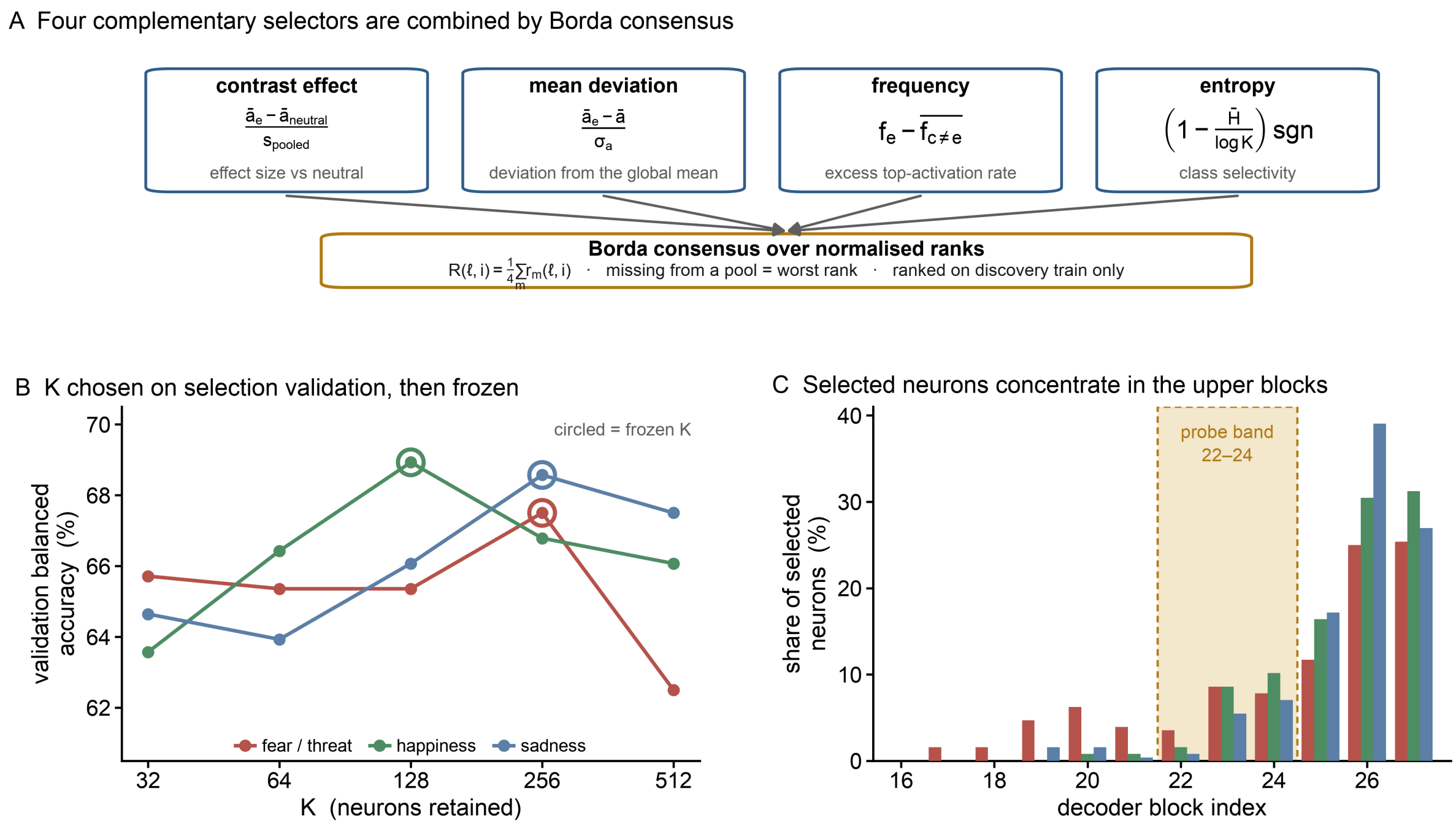

**Figure 7.** Emotion-sensitive neuron localization. (A) The four selectors and the Borda consensus that combines them, all computed on the discovery split alone. (B) Validation balanced accuracy using only the selected neurons, across the five candidate values of $K$; the circled point is the frozen choice for each emotion. (C) Layer distribution of the selected sets, with the probe band outlined; selected neurons concentrate in the upper decoder blocks and extend above the probe band.

## 3.10 Distributed steering directions

The distributed component is a class-mean difference in the residual stream, computed at each candidate layer on the discovery split, following the contrastive-activation construction established for behavioural steering [29,30]:

$$v_e^{(\ell)} = \frac{1}{|\mathcal{D}_e|} \sum_{i \in \mathcal{D}_e} h_i^{(\ell)} - \frac{1}{|\mathcal{D}_{\text{neu}}|} \sum_{i \in \mathcal{D}_{\text{neu}}} h_i^{(\ell)}, \quad \ell \in \{22,23,24\} \quad (18)$$

where $h_i^{(\ell)}$ is the pooled residual-stream state of sample $i$ at the output of block $\ell$ under the frozen token policy.

The choice of reference scale $\rho$ in the steering operator of Eq. (20) matters and the rationale is described explicitly. Setting $\rho = \|v_e^{(\ell)}\|$ makes one unit of $\alpha$ equal to one observed source class-mean contrast, the empirical difference between how the model represents a fear/threat scene and how it represents a neutral one. The alternative, $\rho = \|h^{(\ell)}\|$, makes one unit of $\alpha$ equal to a full residual-stream norm. In this model the mean residual norms at layers 22–24 are approximately 244 to 391 while the direction norms are 33 to 59, so residual-norm scaling would have produced an intervention roughly seven times larger than any contrast present in the data, pushing the hidden state far outside its training distribution and producing degenerate generation rather than a graded behavioural change. We therefore use $\rho = \|v_e^{(\ell)}\|$ throughout, and every reported $\alpha$ is in units of one source class-mean contrast.

The sparse and distributed components are related but are not two parameterisations of one object. The sparse selectors found units across layers 17 to 27, with only 51 of the 256 fear/threat neurons inside layers 22–24, while the directions are defined on layers 22–24 only. Section 4 asks separately of each family whether it changes behaviour.

## 3.11 Intervention operators, hook placement and control construction

Two intervention operators are implemented, shown in Figure 8. The sparse operator multiplies the selected coordinates of the pre-projection activation by a scalar $s$ and leaves every other coordinate bit-identical:

$$\tilde{a}_i^{(\ell)} = s\, a_i^{(\ell)} \text{ for } (\ell, i) \in \mathcal{N}_e, \quad \tilde{a}_i^{(\ell)} = a_i^{(\ell)} \text{ otherwise} \quad (19)$$

with $s = 0$ a knockout, $s \in (0,1)$ an attenuation and $s > 1$ an amplification. It is installed as a forward pre-hook on $W_{\text{down}}^{(\ell)}$. The distributed operator adds a scaled unit direction to the residual stream at the output of block $\ell$:

$$\tilde{h}^{(\ell)} = h^{(\ell)} + \alpha\, \rho\, \frac{v_e^{(\ell)}}{\|v_e^{(\ell)}\|} \quad (20)$$

installed as a forward hook on the decoder block. Both operators are strict no-ops at $s = 1$ and $\alpha = 0$: the hook returns the original tensor object, so instrumented and uninstrumented logits are bit-identical, which we verified before any analysis. A phase tracker distinguishes prefill from decode so that an intervention can be scoped to the prompt, to generation, or to both; all primary conditions use both. Every forward pass checks for non-finite values and aborts any run that produces a NaN.

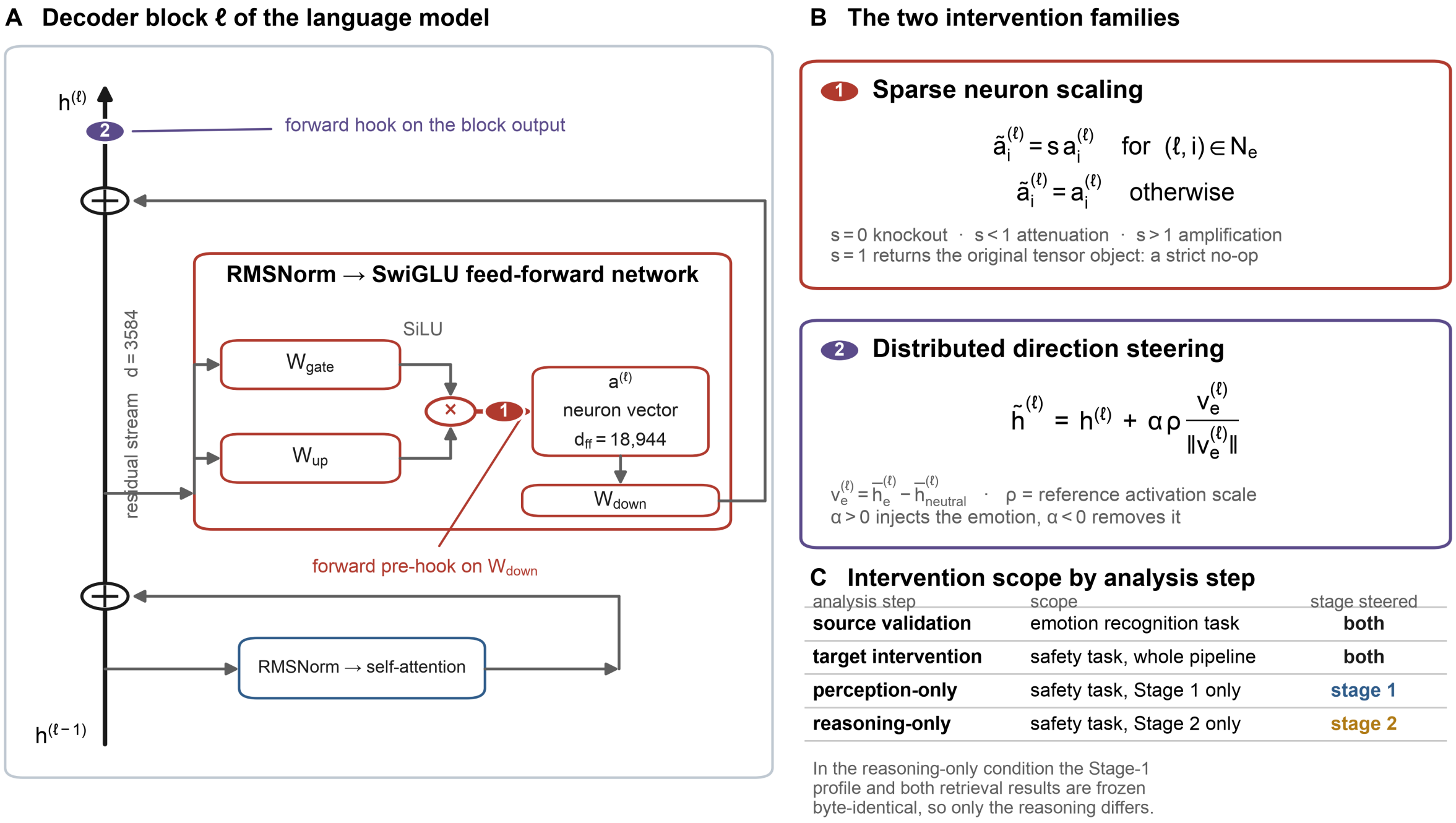

**Figure 8.** Where the two interventions act. (A) One decoder block, with the residual stream, the attention branch and the SwiGLU feed-forward network; hook 1 is a forward pre-hook on the down-projection and hook 2 is a forward hook on the block output. (B) The two intervention families and their operators. (C) The scope at which each experiment applies them; in the reasoning-only condition the Stage-1 profile and both retrieval results are frozen byte-identical across conditions.

Four families of matched control are constructed, each isolating a different alternative explanation.

1. **Same-$K$ random neurons.** Uniformly random $(\ell, i)$ pairs drawn from the same layer universe, testing whether the number of perturbed coordinates is sufficient to produce the effect.
2. **Layer-matched random neurons.** Random coordinates drawn so that the per-layer counts match $\mathcal{N}_e$ exactly, testing whether the depth profile is sufficient.
3. **Activation-magnitude-matched neurons.** For each target neuron, a non-target neuron in the same layer whose mean absolute activation lies within $\pm 10\%$ is drawn without replacement; if no neuron falls inside the tolerance, the nearest-magnitude admissible neuron is used. This preserves both the layer distribution and the total activation energy perturbed, testing whether the effect is simply a function of how much signal is removed.

4. **Shuffled-label circuits.** The class labels of the discovery split are permuted and the entire selection pipeline, including all four selectors, the Borda consensus and the $K$ choice, is re-run on the permuted data. This is the strongest control because it tests the selection procedure itself instead of only its output.

For the distributed family, the corresponding control is a random direction drawn isotropically and rescaled to the norm of $v_e^{(\ell)}$, so that the perturbation magnitude is matched exactly and only its orientation differs. Twenty independent draws are used for each control family in the specificity audit, and a matched-control report records $K$, layer distribution, overlap, Jaccard and mean absolute activation for both target and control so that matching is verifiable rather than asserted.

## 3.12 Source-domain causal validation and operating-point selection

Sparse causal necessity is tested by attenuating the selected neurons at $s \in \{0.75, 0.50, 0.25, 0\}$ and measuring held-out emotion recognition. The quantity of interest is selectivity rather than magnitude, so we define the target-specificity index as the target-class recall drop divided by the mean absolute recall change across the other three classes:

$$\mathrm{TSI}_e = \frac{|\Delta \mathrm{recall}_e|}{\frac{1}{C-1}\sum_{c \neq e} |\Delta \mathrm{recall}_c|} \quad (21)$$

The gate required three conditions simultaneously: $\mathrm{TSI}_e > 1$, a target recall drop of at least 0.05, and an effect larger than the best matched control. Three matched draws per emotion supported the gate; the larger audit in Section 4.7 uses 20 draws per family.

Behavioural sufficiency is tested over sparse scales $s \in \{0, 0.5, 0.75, 1.25, 1.5, 2\}$ and distributed strengths $\alpha \in \{-2, -1, -0.5, 0.5, 1, 2\}$, with $s = 1$ and $\alpha = 0$ supplied by the baseline. An operating point is usable only if the intervention leaves the model functional, which we enforce with an explicit non-degeneracy rule: overall balanced accuracy must remain at least 85% of baseline with an absolute floor of 0.519, and the target class must not exceed 80% of all predictions. This rule rejects saturation, because a strength that drives the model to answer “fear” to everything would otherwise score well on target-class tendency while being behaviourally useless. Fear/threat passed at $\alpha = \pm 0.5$, happiness at $\alpha = +0.5$, and sadness at no tested distributed strength, which is itself reported as a finding in Section 4.7.

## 3.13 Target intervention, stage resolution and circuit analysis

The central target experiment applies the frozen distributed fear/threat direction to all 65 building cases during both stages. Seven conditions are compared: baseline; positive steering at $\alpha = +0.5$; negative steering at $\alpha = -0.5$; a norm-matched random direction at $\alpha = +0.5$; fear-neuron amplification at $s = 2$; fear-neuron knockout at $s = 0$; and a layer-matched random-neuron control. Cases, prompts, retrieval, rules, token budgets, candidate strings and decoding are identical across paired conditions, and eight independently generated baselines produced identical predictions on all 65 cases, confirming determinism.

Happiness at $\alpha = +0.5$ is the matched positive-valence control. Sadness was intended as a negative-valence, lower-threat control, but no distributed sadness strength passed the source usability rule, so sadness is compared only within the sparse family. The design can therefore reject the claim that any affective direction, or positive valence specifically, reproduces the fear result.

For perception-only intervention, the direction is active while Stage 1 generates the profile and is removed before retrieval and Stage 2. Effects are measured on the final posting, on each parsed field, on hazard severities and on schema compliance. For reasoning-only intervention, every case reuses the baseline Stage-1 text, parsed profile, retrieved cases, retrieved rules and complete Stage-2 context, with byte-identity verified case by case through text and context hashes.

The circuit comparison localizes two structural sparse sets on 200 randomly selected train-fit cases under seed 37. Activations are recorded at the Stage-2 decision position with the published ground-truth Stage-1 profile supplied, so that perception is held constant and the sets reflect the decision computation rather than upstream perception. An unsafe-decision set contrasts Red against non-Red ground truth, and a damage-severity set contrasts the highest against the lowest observed maximum hazard severity. Both use the contrast and mean-deviation selectors, Borda aggregation, the same $K$, and the same $3 \times 18{,}944 = 56{,}832$-neuron layer-band universe as the emotion sets, so overlaps are computed between comparable objects. Overlap is quantified by Jaccard index and tested against a hypergeometric null.

## 3.14 Statistical analysis

Every target comparison is paired at case level. Confidence intervals for changes in macro-F1, balanced accuracy and Red recall come from a paired case-level bootstrap with 10,000 resamples, in which both conditions are recomputed on the same resampled case indices so that the pairing is preserved. A cluster bootstrap resampling whole earthquake events, with 1,000 resamples, is reported as a robustness analysis, because cases from one event are not independent. McNemar's exact test evaluates paired correctness. For control families, the empirical null $p$-value uses the conservative estimator

$$p = \frac{1 + |\{j: \delta_j \geq \delta_{\text{obs}}\}|}{1 + n_{\text{draws}}} \quad (22)$$

which cannot return zero. Benjamini–Hochberg correction controls the false discovery rate at 0.05 within the family of 24 macro-F1 steering comparisons. Seed 37 governs all splitting, sampling and control construction.

# 4. Results

## 4.1 Evidence of a badly placed criterion

We first examine whether the target pipeline exhibits the defect the method is designed to address. The results show that it does, and that the two SDT components separate cleanly. Table 1 reports the three Stage-1 conditions on the test split.

**Table 1.** Structural-safety testbed on the published SeisMLLM-1K test split ($n = 65$). The primary few-shot condition retains moderate Red-versus-rest discrimination while placing the criterion far into the evidence-demanding region.

| Stage-1 condition | Accuracy [95% CI] | Macro-F1 | Balanced accuracy | Red recall | Red FNR | $d'$ | $c$ |
|---|---|---|---|---|---|---|---|
| Oracle published profile | 0.692 [0.585, 0.800] | 0.667 | 0.724 | 0.622 | 0.378 | 2.416 | +0.907 |
| Few-shot (primary) | 0.400 [0.277, 0.523] | 0.382 | 0.447 | 0.270 | 0.730 | 1.521 | +1.354 |
| Zero-shot | 0.262 [0.154, 0.369] | 0.138 | 0.333 | 0.000 | 1.000 | −0.107 | +2.168 |

The primary few-shot reconstruction reached 0.400 accuracy (95% CI [0.277, 0.523]), macro-F1 0.382 and balanced accuracy 0.447. It predicted 10 Red cases against 37 true Red cases, giving Red recall 0.270 and a Red false-negative rate of 0.730. Red precision, however, was 1.000: every Red posting it issued was correct. A model that is never wrong when it says Unsafe and yet misses nearly three-quarters of unsafe buildings indicates not a failure to perceive, but a reluctance to act on what it perceives. The SDT decomposition corroborates this directly: $d' = 1.521$ with $c = +1.354$. In operational terms this is extreme permissiveness about occupancy, since everything short of unambiguous collapse is passed as Green or Yellow. The pattern matches what has been reported for multimodal models in adjacent safety tasks, where zero-shot hazard recall falls below 0.33 and a substantial fraction of real hazards goes unflagged [7,12].

Three further observations establish that this is a property of the decision threshold. First, supplying the published ground-truth Stage-1 profile raised accuracy to 0.692 and $d'$ to 2.416 while $c$ remained high at +0.907, so better perception improves discrimination without by itself relocating the criterion. Second, parse failure does not explain the baseline: accuracy was 0.394 among the 33 cases with a cleanly parsed profile and 0.406 among the remaining 32. Third, the pattern is systematic across

classes: per-class precision/recall/F1 were 0.343/0.706/0.462 for Green, 0.200/0.364/0.258 for Yellow and 1.000/0.270/0.426 for Red, with predicted counts 35/20/10 against a true distribution of 17/11/37.

Figure 9 makes the consequences of that criterion placement more explicit, and in doing so separates two questions that the reported metrics conflate. Panel A shows the mismatch directly: the model issues 35 Green postings where 17 are warranted and 10 Red postings where 37 are warranted, so 27 unsafe buildings are passed as occupiable or merely restricted. Panel B sweeps a decision threshold over the model's own candidate probability $\hat{p}(\text{Red})$ from Eq. (9) and tracks the reported metrics as the number of Red postings grows. Every quantity an engineering evaluation would normally report moves substantially along this curve while the underlying evidence is untouched: Red recall rises monotonically from 0.27 to 1.00, Red precision falls from 1.00, and macro-F1 traces an inverted U that peaks at 0.453 when 32 Red postings are issued. Reporting any single point on this curve therefore describes a policy choice, not a competence.

As shown in Panel C, driven by external thresholding to exactly the 38 Red postings that the frozen fear direction produces in Section 4.4, the baseline attains macro-F1 0.376 and $d' = 0.529$; the internal intervention at the same count attains 0.556 and 1.027. The two routes reach almost the same criterion ($c = -0.178$ and $c = -0.161$) by different means, and they are not equivalent: thresholding the output distribution of a poorly calibrated model degrades discrimination far more than steering its internal representation does. The best macro-F1 achievable by thresholding alone, 0.453, remains below the steered value.

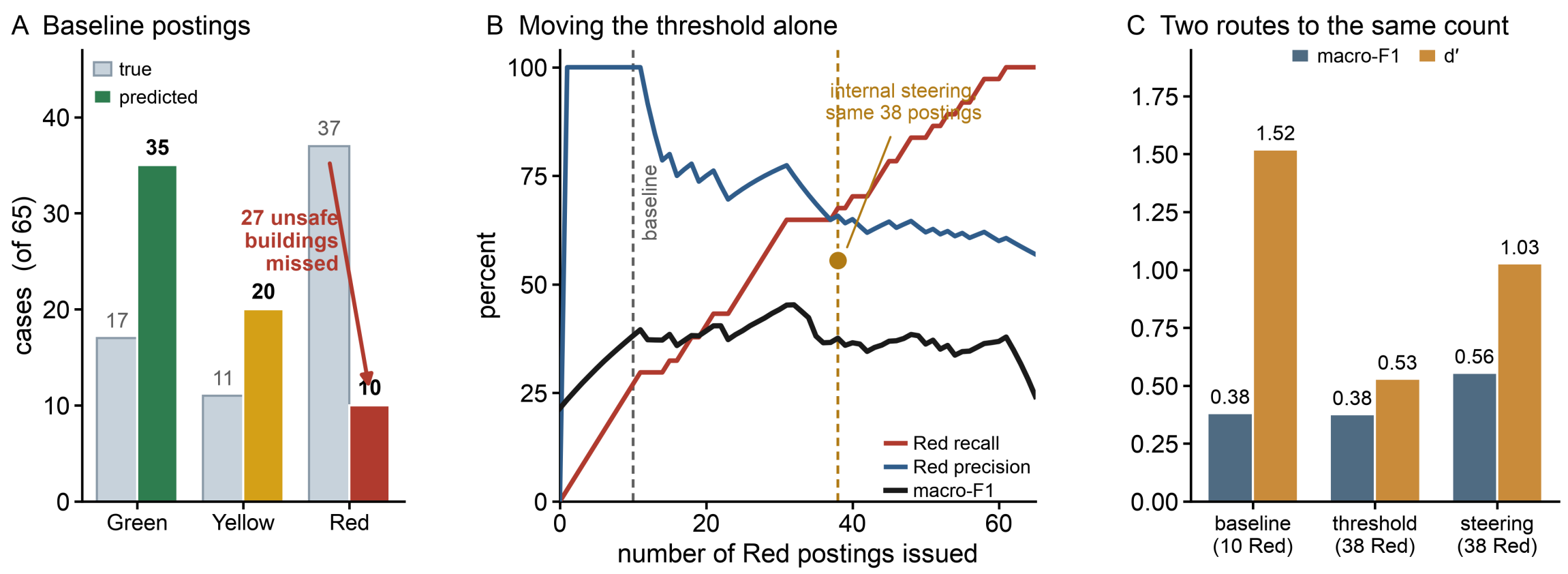

**Figure 9.** How criterion placement alone moves the reported metrics. (A) Baseline predicted postings against the true distribution; 27 of 37 unsafe buildings are missed. (B) Sweeping a decision threshold over the model's own candidate probability $\hat{p}(\text{Red})$: recall, precision and macro-F1 all move substantially while the evidence is unchanged. Dashed lines mark the baseline operating point (10 Red postings) and the count produced by internal steering (38). (C) The two routes to 38 Red postings compared: external thresholding of the baseline's probabilities against the internal fear direction of Section 4.4.

The accuracy gap of our reproduced SeisMLLM locates most of the reconstruction deficit in Stage-1 perception, which the published system fine-tunes and this study does not. Calibration degraded correspondingly, with the multiclass Brier score rising from 0.416 under oracle perception to 0.926 under few-shot and expected calibration error from 0.132 to 0.334. Despite weaker absolute performance, the testbed supplies exactly what our method aims to address: real but under-used discriminative evidence together with a criterion in the wrong place, so that criterion movement is both measurable and worth making.

## 4.2 Affect is linearly decodable in late layers and localizable to sparse neurons

Before any transfer, we must be able to localize the affect representation. Under the leakage-resistant last-prompt-token policy, held-out four-class balanced accuracy rose from 0.250 at layer 0 (exactly chance, as it must be if the policy is clean) to 0.714 (95% CI [0.660, 0.767]) at layer 22, and validation selected layers 22–24 as the candidate band (Figure 10). Best-layer recall was 0.700 for neutral, 0.800 for happiness, 0.729 for sadness and 0.629 for fear/threat. Valence peaked at $R^2 = 0.729$ (Pearson $r = 0.854$) at layer 22, whereas arousal peaked at only $R^2 = 0.297$ ($r = 0.551$) at layer 26. Affect is thus available in late residual representations, with valence substantially more linearly accessible than arousal, and the two peaking at different depths. This is consistent with the linear encoding of sentiment reported for text-only models [31] and with the late-layer emergence of semantically explicit visual content in VLMs [27].

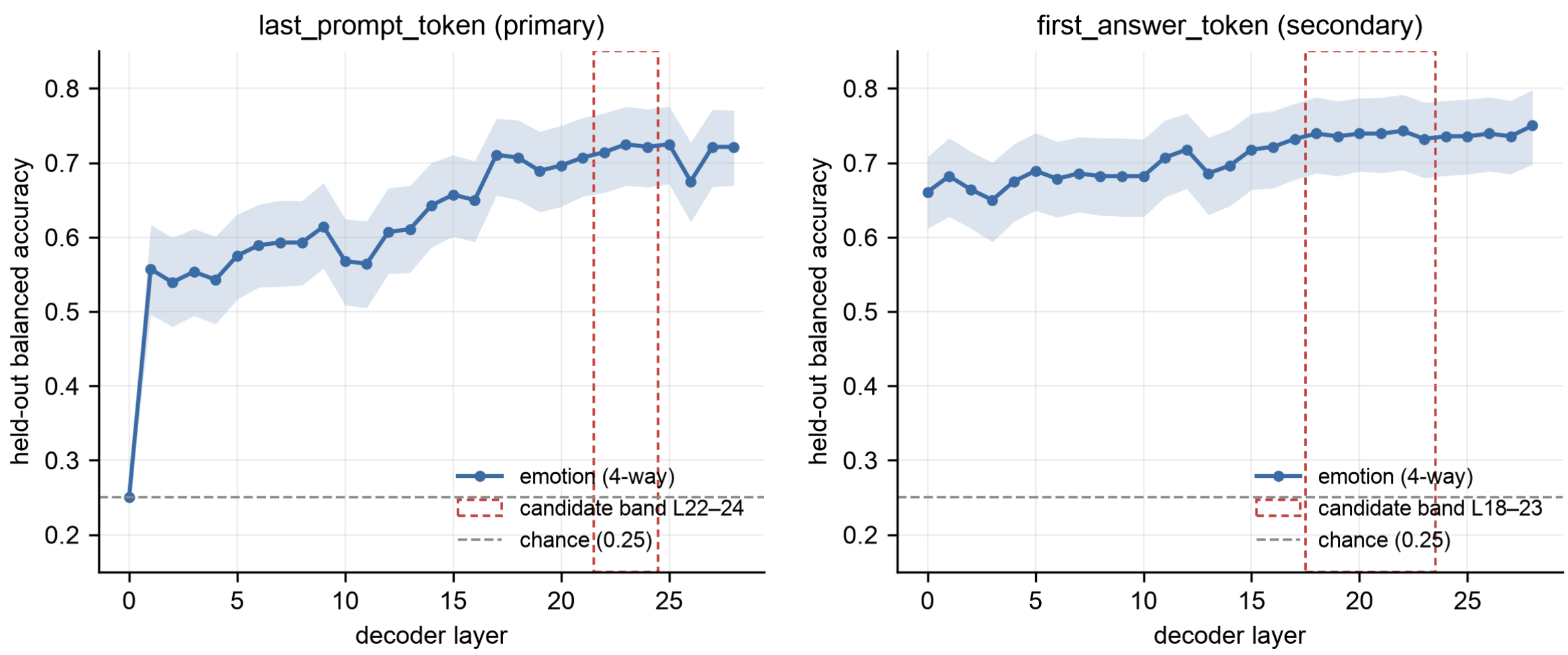

**Figure 10.** Layer-wise affect representation. Held-out emotion decoding at the last prompt token emerges with depth and peaks at layer 22; validation selected layers 22–24 as the candidate band.

Sparse selection produced 256 fear/threat, 128 happiness and 256 sadness neurons, distributed across late layers (Figure 11). Validation performance was insensitive to $K$ within the tested range (0.654 to 0.675 for fear, 0.636 to 0.689 for happiness and 0.639 to 0.686 for sadness), which is why $K$ was frozen on validation rather than tuned further. Individual selectors agreed poorly, with pairwise

Jaccard between 0.000 and 0.030, confirming that the four criteria of Section 3.9 capture genuinely different senses of emotion sensitivity. Consensus sets, however, were moderately stable under resampling: fear Jaccard 0.588 ± 0.033 against the full-sample set, with 123 of 256 neurons present in at least 80% of bootstrap resamples, and 0.678 and 0.682 for happiness and sadness. Five-fold agreement for fear was 0.203. The results support a reproducible late-layer sparse component, but to avoid treating any single ranking rule as the definition of the circuit, we further conducted causal validation.

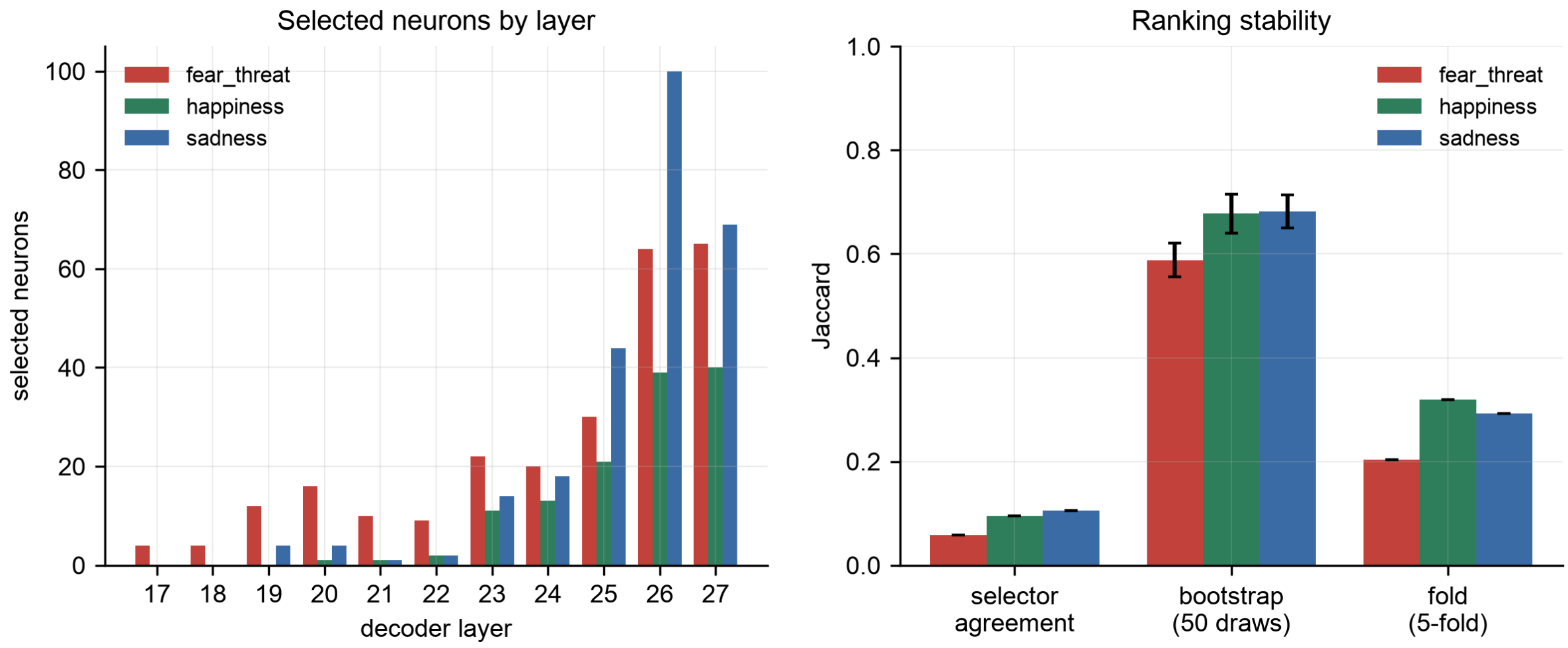

**Figure 11.** Sparse circuit localization and stability. Selected affect-sensitive neurons concentrate in late decoder layers. Bootstrap consensus is substantially more stable than agreement among individual selectors or across folds, motivating ensemble selection followed by causal validation.

## 4.3 The source circuit is causally necessary and behaviourally sufficient

As decodability is not causality, we further tested the identified circuit with intervention experiments on held-out source data. The unintervened held-out source task had balanced accuracy 0.657 and macro-F1 0.641. Full sparse knockout reduced fear/threat recall by 0.129 while the mean recall of the other three classes changed by −0.005, giving a target-specificity index of 27.0 (Eq. 21). Corresponding target-recall drops were 0.086 for happiness and 0.057 for sadness, with specificity indices 18.0 and 3.0 (Table 2, Figure 12). Matched full-knockout controls remained at approximately zero. Most of the impairment appeared only near complete knockout, and the knockout strength did not show a linear effect.

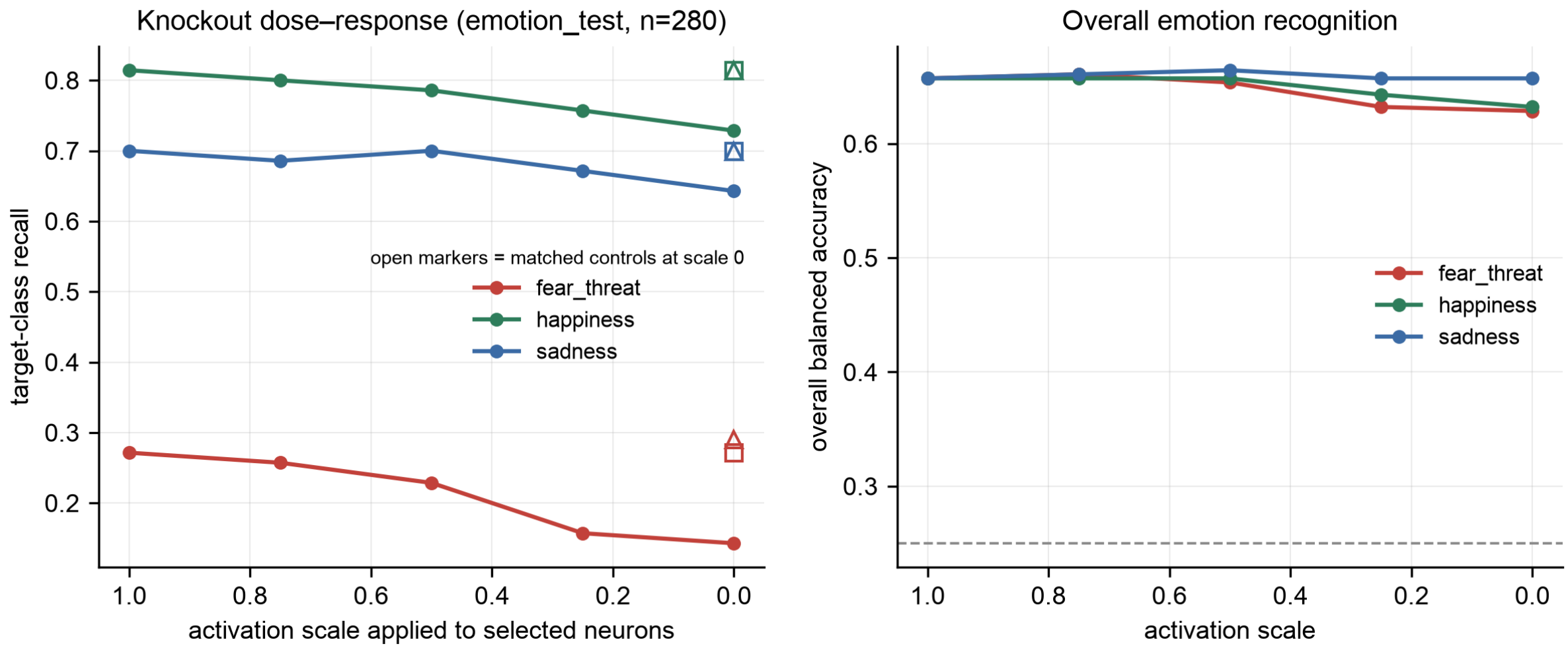

**Figure 12.** Sparse circuit knockout on the source task. Scaling the selected neurons toward zero selectively reduces the corresponding emotion's recall while overall balanced accuracy changes far less. Open markers show matched controls at full knockout, which cluster at zero.

**Table 2.** Source-domain causal evidence on the held-out emotion-test split ($n = 280$). Sadness passes the sparse necessity gate but admits no usable distributed strength, which blocks the intended distributed valence contrast.

| Evidence | Fear/threat | Happiness | Sadness | Interpretation |
|---|---|---|---|---|
| Sparse neurons, $K$ | 256 | 128 | 256 | Validation-selected set size |
| Full-knockout target-recall drop | 0.129 | 0.086 | 0.057 | Source-task causal necessity |
| Other-class mean recall change | −0.005 | +0.005 | −0.019 | Selectivity of the effect |
| Target-specificity index (Eq. 21) | 27.0 | 18.0 | 3.0 | All three passed the source gate |
| Usable distributed $\alpha$ | −0.5, +0.5 | +0.5 | none | Non-degenerate steering only |

Distributed steering produced graded source behaviour only within a narrow range, which is why the non-degeneracy rule of Section 3.12 matters (Figure 13). On held-out images, fear at $\alpha = +0.5$ increased the predicted fear fraction by 0.457, fear at $\alpha = -0.5$ reduced it by 0.079, and a norm-matched random direction increased it by 0.086. On validation, fear predictions rose from 0.054 at baseline to 0.129 at $\alpha = +0.5$, 0.179 at $+1.0$ and 0.604 at $+2.0$, but balanced accuracy collapsed from 0.611 to 0.339 and 0.368 at the two larger strengths. Every positive sadness direction saturated at 0.986 to 1.000 sadness predictions with balanced accuracy of 0.250 to 0.264, that is, at chance. These conditions were rejected as degenerate, and the modest $\alpha = \pm 0.5$ was consequently selected as the frozen operating point for fear.

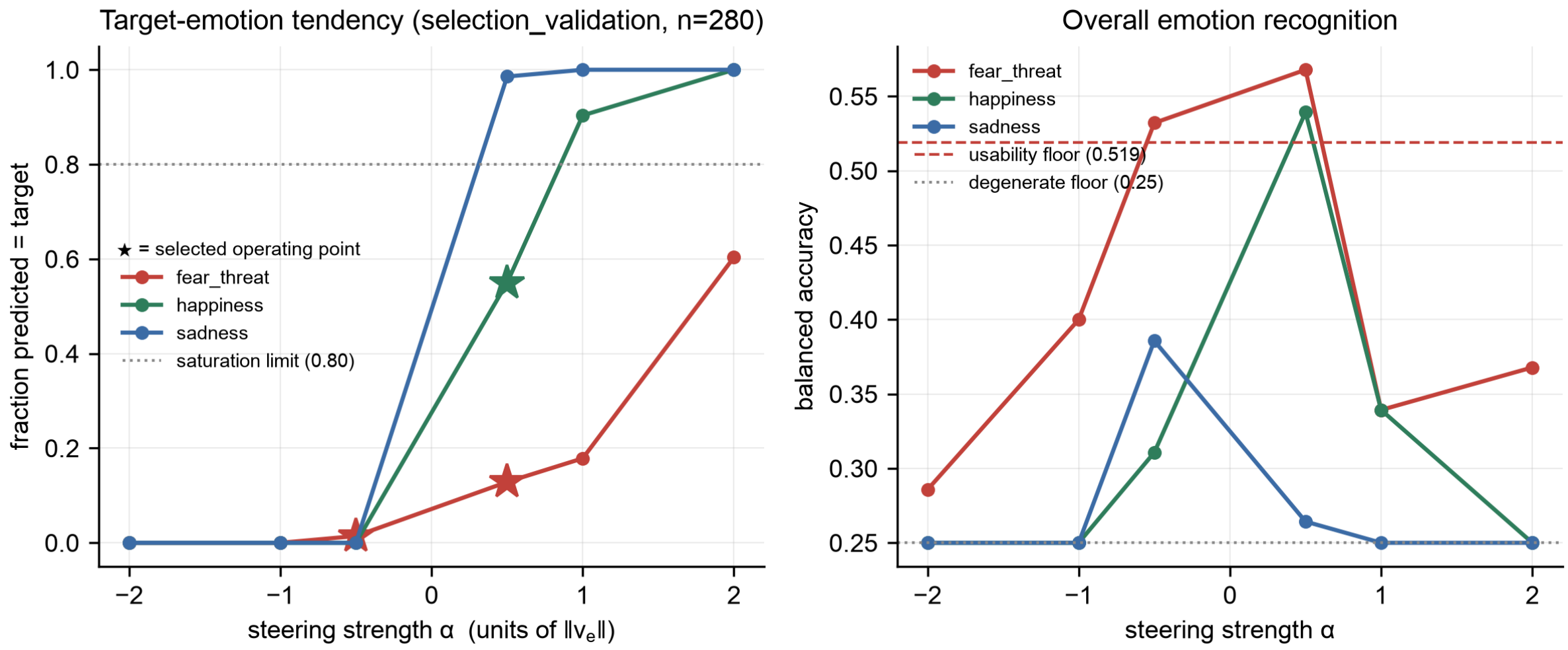

**Figure 13.** Distributed source steering and operating-point selection. Target-class tendency and overall source-task balanced accuracy are plotted against intervention strength. Stars mark the non-degenerate strengths frozen for transfer. The large fear strengths and all positive sadness strengths failed the usability rule and were discarded.

To summarise, emergence with depth, selective knockout, four null control families and usable bidirectional steering together establish a causally manipulable affective circuit on the independent source task.

## 4.4 The frozen circuit moves the building decision bidirectionally

With the identified circuit frozen, this is the test of our proposed method on the target dataset. Positive distributed fear/threat steering raised Red recall from 0.270 to 0.757, a paired change of +0.486 (95% CI [+0.306, +0.667], $p < 0.001$). Macro-F1 increased by 0.174 (95% CI [+0.014, +0.331], $p = 0.034$), and predicted counts moved from 35/20/10 Green/Yellow/Red to 13/14/38, close to the true distribution of 17/11/37. Subtracting the same direction was directionally symmetric: Red recall fell to 0.000, a change of −0.270 (95% CI [−0.419, −0.133], $p < 0.001$), macro-F1 decreased by 0.187, and predictions became 50/15/0 (Figure 14). These findings therefore confirm that the identified affective circuit can control the Unsafe boundary in the building damage assessment pipeline in both directions, with no weight updated and no prompt changed.

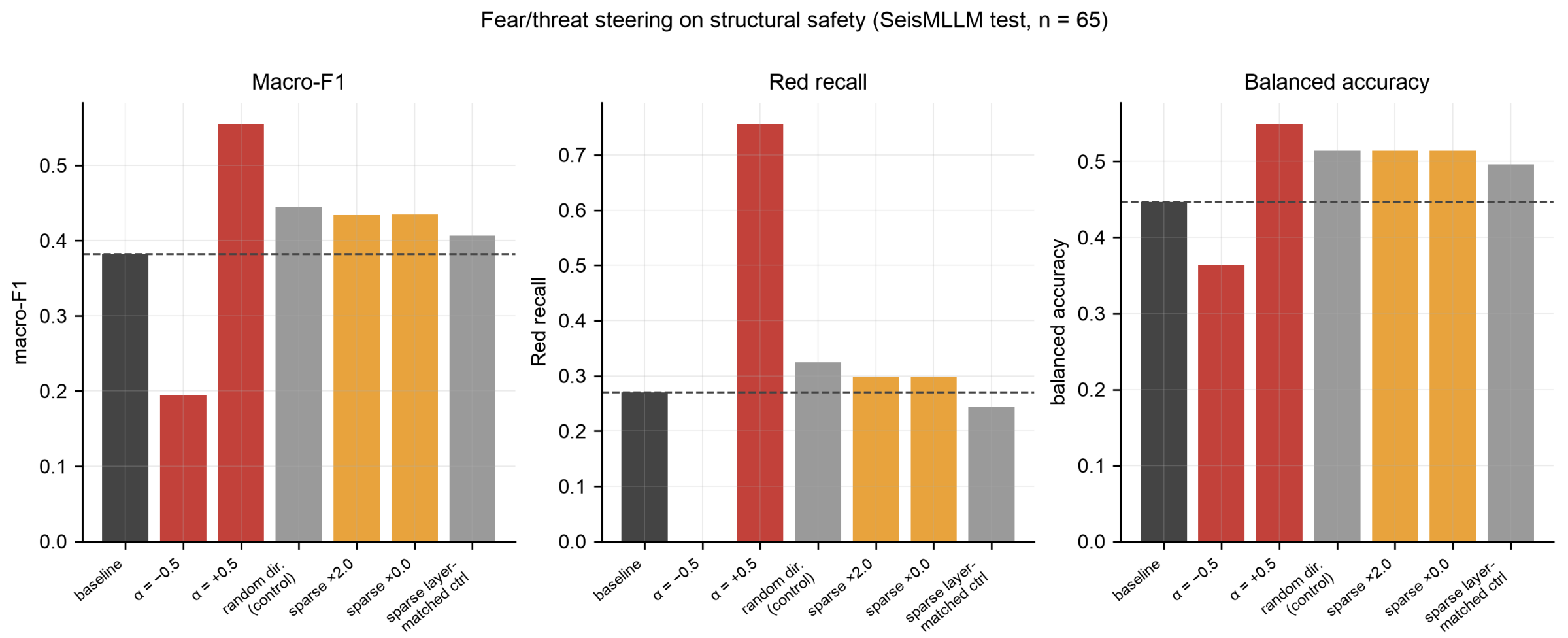

**Figure 14.** Distributed fear/threat intervention on building safety. Positive and negative steering move macro-F1, Red recall and predicted class counts in opposite directions. Neither the norm-matched random direction nor the sparse interventions reproduces the increase in Red decisions.

The control experiments help establish the causal intervention. Table 3 collects every target condition with its paired effect sizes. The norm-matched random direction at $\alpha = +0.5$, which perturbed the hidden state by the same magnitude but along an arbitrary orientation, increased Red recall by only +0.054 (95% CI [−0.051, +0.162], $p = 0.457$) and macro-F1 by +0.064 ($p = 0.222$), with predicted Green/Yellow/Red counts of 30/21/14. Because the confidence interval for Red recall includes zero, perturbation magnitude alone does not reliably reproduce the effect, indicating that the orientation of the intervention in representation space is what matters. Likewise, sparse fear amplification and knockout each changed Red recall by +0.027, indistinguishable from the layer-matched sparse control. As a result, the behavioural change cannot be attributed simply to increasing or suppressing a small set of individual units. Nor can it be attributed to generation degeneracy, since Stage-1 token-cap rates were 1.5% to 4.6% across reported conditions against 1.5% at baseline.

**Table 3.** FearCaut-Qwen target effects on the published test split ($n = 65$ paired). Baseline predicted counts were 35/20/10 against a true distribution of 17/11/37. Confidence intervals and $p$-values are from a paired case-level bootstrap with 10,000 resamples.

| Condition | Δ macro-F1 [95% CI] | $p$ | Δ Red recall [95% CI] | $p$ | $\Delta d'$ | $\Delta c$ | G/Y/R predicted |
|---|---|---|---|---|---|---|---|
| Fear direction, $\alpha = +0.5$ | +0.174 [+0.01, +0.33] | 0.034 | +0.486 [+0.31, +0.67] | <0.001 | −0.493 | −1.515 | 13/14/38 |
| Fear direction, $\alpha = -0.5$ | −0.187 [−0.31, −0.05] | 0.006 | −0.270 [−0.42, −0.13] | <0.001 | −1.628 | +0.814 | 50/15/0 |
| Norm-matched random direction, | +0.064 [−0.04, +0.17] | 0.222 | +0.054 [−0.05, +0.16] | 0.457 | −0.599 | −0.450 | 30/21/14 |
| $\alpha = +0.5$ | | | | | | | |
| Fear sparse amplification, $s = 2$ | +0.052 [−0.04, +0.15] | 0.270 | +0.027 [−0.13, +0.19] | 0.873 | −0.409 | −0.281 | 37/16/12 |
| Fear sparse knockout, $s = 0$ | +0.053 [−0.03, +0.14] | 0.235 | +0.027 [−0.06, +0.12] | 0.809 | −0.409 | −0.281 | 34/19/12 |
| Sparse layer-matched control | +0.025 [−0.08, +0.13] | 0.633 | −0.027 [−0.15, +0.09] | 0.829 | −0.081 | +0.040 | 42/14/9 |

Positive steering changed 40 of 65 postings. Thirty-eight moved toward greater severity (15 Green→Red, 14 Yellow→Red and 9 Green→Yellow), while two moved the other way. As shown in Figure 15, positive fear/threat steering converts many baseline Green and Yellow predictions to Red, including cases whose true label is not Red, and negative steering eliminates Red predictions entirely, whereas the random and sparse controls produce smaller or qualitatively different changes. These results show that the intervention does not merely repair baseline false negatives: it also converts genuinely non-Red cases to Red. That is the signature of a moved threshold rather than of better evidence, which is further confirmed quantitatively in Section 4.5.

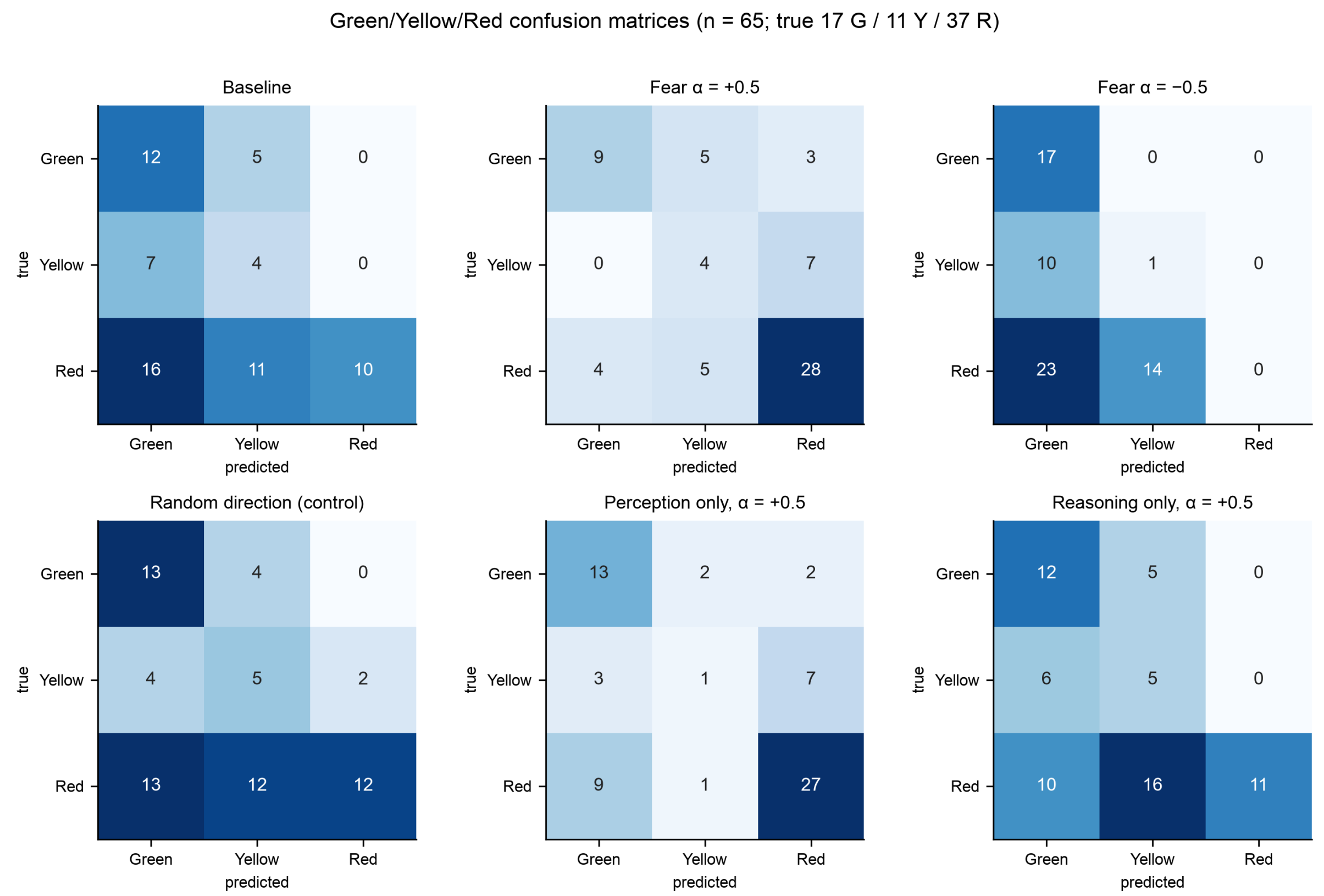

**Figure 15.** Confusion matrices under the key target conditions.

### 4.5 The effect is criterion movement, and discrimination is not improved

Positive fear/threat steering moved the criterion from $c = +1.354$ to $c = -0.161$, a shift of $\Delta c = -1.515$, while discrimination fell from $d' = 1.521$ to $d' = 1.028$, $\Delta d' = -0.493$. The criterion-to-discrimination ratio of Eq. (4) is $\rho_{c/d} = 3.1$: three-quarters of the behavioural change is threshold movement, and none of it is improved discrimination (see Figure 16).

This shows the intended result and the designed capability. FearCaut-Qwen is a criterion lever, and it moves the criterion by one and a half units of standardised evidence, from a strongly evidence-demanding position to one almost exactly at the midpoint between the unsafe and non-unsafe evidence distributions. Whether that new position is the right one for a given deployment is an engineering decision about the cost of a missed hazard relative to the cost of an unnecessary closure, which is not totally within the scope of this study. Our results settle the mechanism that a decision criterion can be shifted by affection steering.

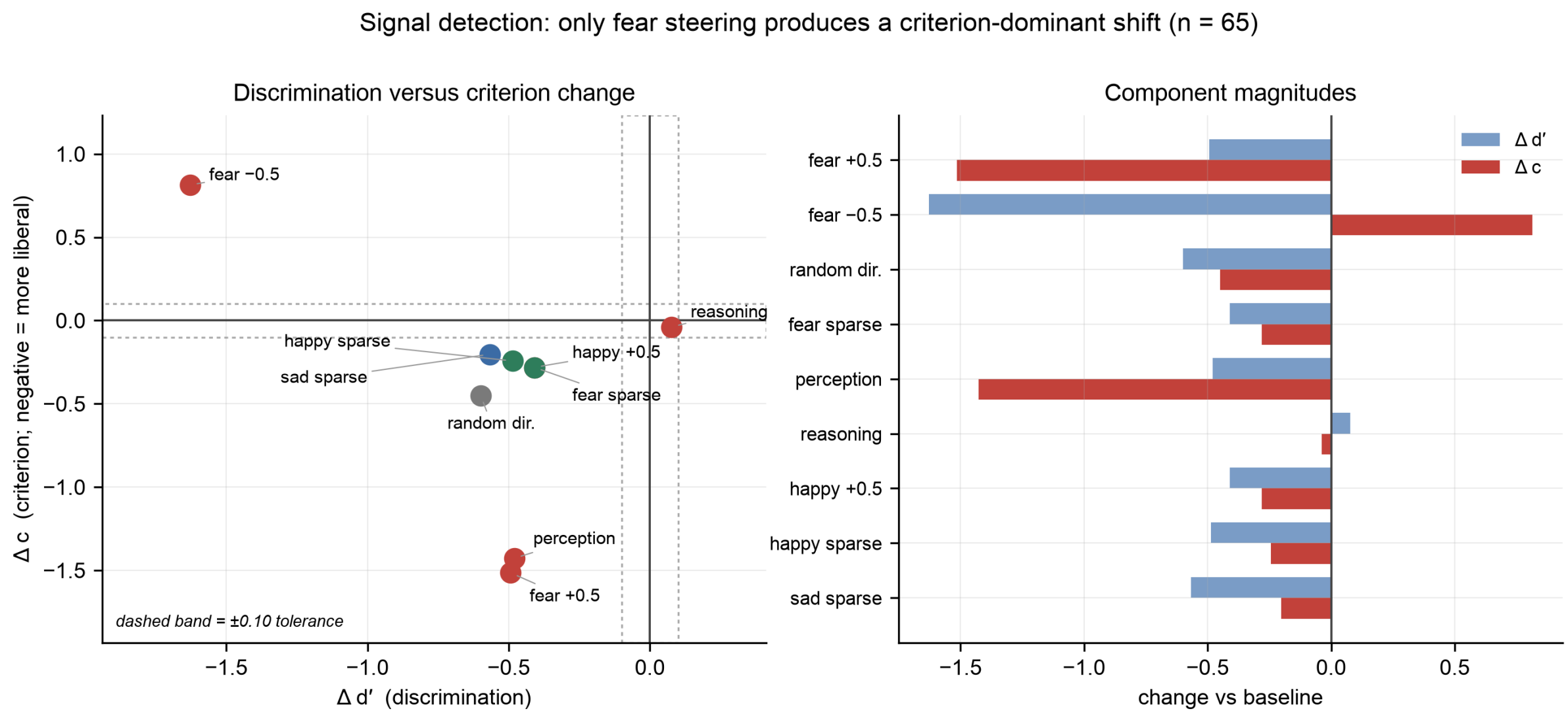

**Figure 16.** SDT decomposition across interventions. Left: each condition positioned by its change in discrimination and in criterion. Right: component magnitudes. Positive distributed fear/threat steering is dominated by a liberal criterion shift; no tested intervention produced a positive $\Delta d'$ beyond the 0.10 tolerance.

The remaining conditions help separate a change in decision policy from a general degradation of the representation. Negative steering raised the criterion ($\Delta c = +0.814$) and also reduced discrimination substantially ($\Delta d' = -1.628$), eliminating Red predictions. The matched random direction reduced discrimination ($\Delta d' = -0.599$) while moving the criterion much less ($\Delta c = -0.450$). This suggests that a perturbation of similar magnitude can disrupt the quality of the internal representation without necessarily shifting the model's decision boundary. The sparse layer-matched control produced little change in either component ($\Delta d' = -0.081$, $\Delta c = +0.040$), indicating that manipulating a comparable set of units in the same layers is not sufficient to reproduce the effect.

It is worth emphasising that classification metrics can be improved by moving the decision threshold without making the model better at understanding building damage, although this is not intuitive. First, regarding the macro-F1 gain of +0.174: the test set contains 37 Red cases out of 65, whereas the baseline predicts Red only rarely because its decision criterion is strongly conservative, so when the intervention shifts that criterion toward a more appropriate position, more true Red cases are identified and macro-F1 increases mechanically. The quality of the underlying evidence is not changed. Consistently, only one of the 24 macro-F1 comparisons survives Benjamini–Hochberg correction, and the threshold sweep of Figure 9(B) shows that a comparable macro-F1 movement is obtainable from the baseline scores alone. This flags again the importance of the SDT decomposition, since any evaluation reading accuracy alone would have recorded this intervention as an improvement in structural understanding. Second, none of the tested conditions produced a meaningful increase of $\Delta d'$ beyond the 0.10 tolerance, indicating that the intervention did not make unsafe and non-unsafe buildings more separable in the model's underlying evidence, but acted as a representational shift in decision policy.

## 4.6 Criterion control originates during perception, not during rule-based reasoning

This subsection asks where in the two-stage pipeline the threshold movement occurs. The stage-resolved experiments are able to answer this cleanly, because the reasoning-only condition holds the evidence and the retrieved context byte-identical.

As shown in Table 4 and Figure 17, perception-only steering increased Red recall by 0.459 (95% CI [+0.278, +0.634], $p < 0.001$), recovering 94.4% of the whole-pipeline effect, with $\Delta d' = -0.480$ and $\Delta c = -1.428$ closely matching the both-stage condition; macro-F1 increased by 0.116 ($p = 0.083$). Reasoning-only steering, however, increased Red recall by only 0.027 (95% CI [+0.000, +0.091], $p = 0.743$), 5.6% of the whole effect, with $\Delta d' = +0.077$ and $\Delta c = -0.038$, both inside the no-material-change tolerance.

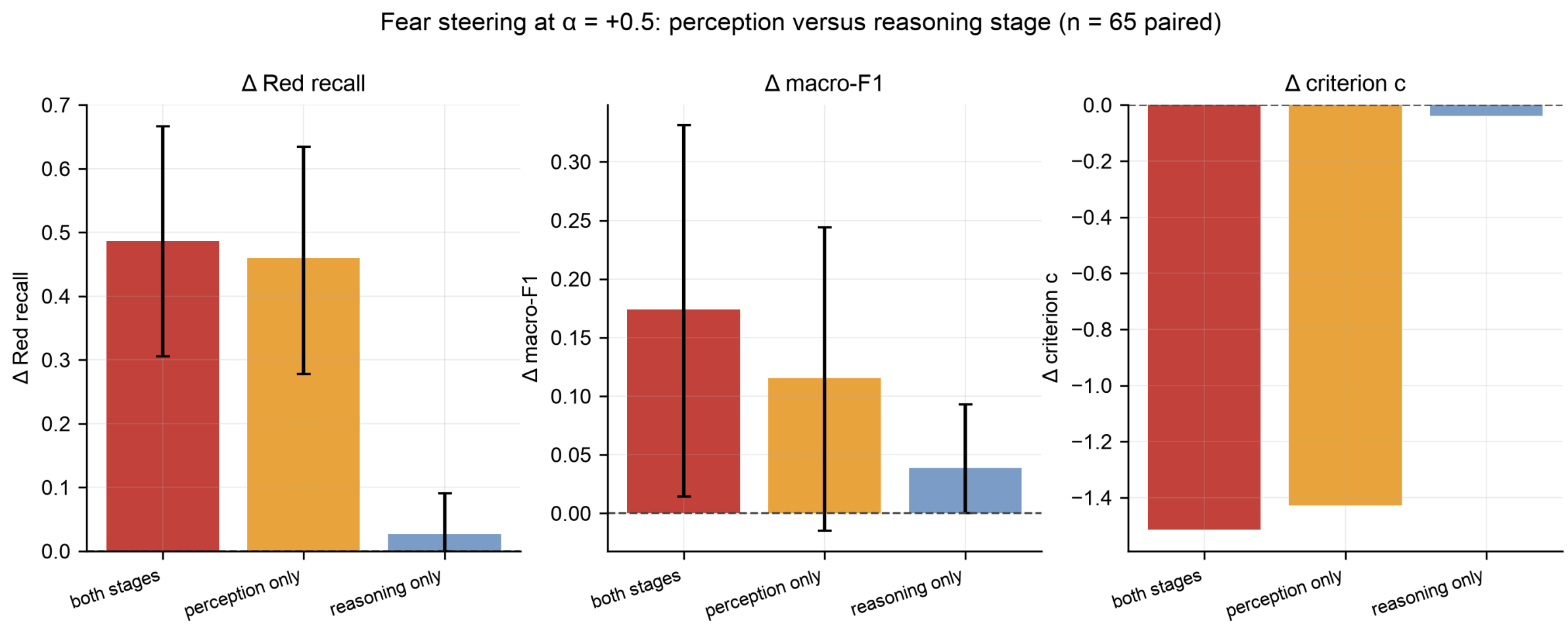

**Figure 17.** Perception-versus-reasoning stage decomposition. Perception-only intervention recovers nearly the entire Red-recall and criterion effect. Reasoning-only intervention over exactly fixed evidence and retrieved context is small and not significant.

**Table 4.** Stage-resolved intervention effects for fear/threat at $\alpha = +0.5$. Shares are descriptive proportions of the absolute Red-recall effect.

| **Intervention stage** | **Δ Red recall [95% CI]** | $p$ | **Share of full effect** | **Δ macro-F1** | $p$ | $\Delta d'$ | $\Delta c$ |
|---|---|---|---|---|---|---|---|
| Both stages | +0.486 [+0.306, +0.667] | <0.001 | 100.0% | +0.174 | 0.034 | −0.493 | −1.515 |
| Perception only | +0.459 [+0.278, +0.634] | <0.001 | 94.4% | +0.116 | 0.083 | −0.480 | −1.428 |
| Reasoning only | +0.027 [+0.000, +0.091] | 0.743 | 5.6% | +0.039 | 0.047 | +0.077 | −0.038 |

The Stage-1 perceptual reports changed substantially under steering, but the changes were not a simple increase or decrease in perceived damage severity. Under positive steering, mean summed hazard-severity rank rose from 5.02 to 6.57 (+1.55), with 21 cases more severe and 15 less severe. However, negative steering also raised the mean profile rank, from 5.31 to 6.60 (+1.29), while moving the final decision in the opposite direction. Among the 47 cases whose baseline and steered profiles both parsed, the correlation between severity change and ordinal decision change was only +0.308. Construction-type agreement fell to 27 of 65, and Stage-1 parse success remained similar across conditions (0.508 at baseline, 0.462 under positive and 0.538 under negative steering). The intervention therefore acts during multimodal evidence construction, not through a simple monotonic change in perceived damage. In other words, what changes is the whole character of the perceptual report, and the decision follows.

## 4.7 The effect is threat-specific

Specificity is what separates a criterion lever from a generic perturbation. At the same usable strength $\alpha = +0.5$, the happiness direction changed Red recall by +0.027 ($p = 0.84$) and macro-F1 by +0.006 ($p = 0.94$), against fear/threat changes of +0.486 and +0.174 (Figure 18). A positive-valence affective direction of the same construction and magnitude does not reproduce the effect, which rules out an affect-general account and is consistent with the emotion-specificity of appraisal-tendency effects in humans [20,21]. In the sparse family, fear, happiness and sadness changed Red recall by +0.027, +0.000 and −0.027 respectively, all null. Because no sadness distributed strength passed the source usability criterion, the study cannot complete a matched distributed comparison between two negative-valence classes.

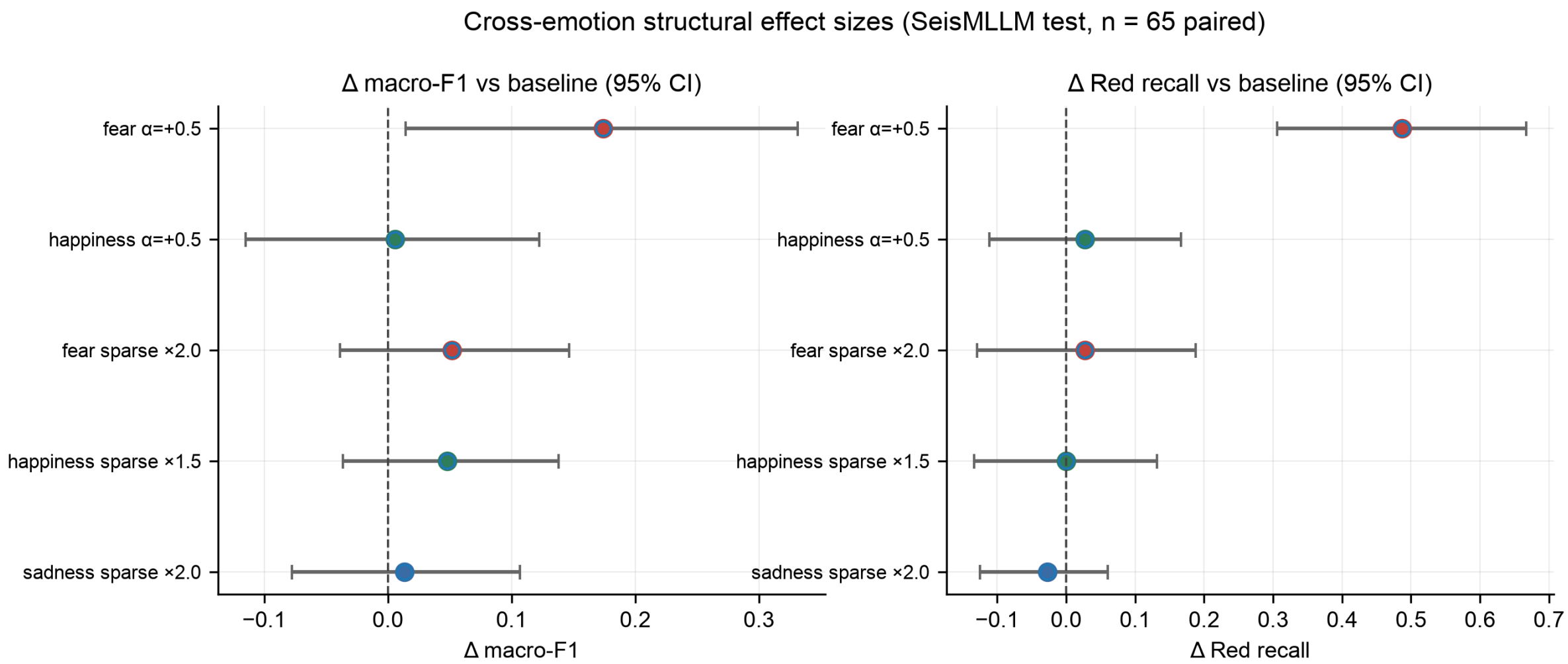

**Figure 18.** Cross-emotion target effects. Fear/threat and happiness are compared at the same distributed strength $\alpha = +0.5$ ; only fear/threat produces the large Red-recall shift. Sparse interventions are null for all three emotions. No valid distributed sadness point exists, so fear/threat cannot be fully separated from generic negative valence.

Source specificity was evaluated using four null-control families, each sampled independently 20 times (Figure 19). The purpose of these controls was to determine whether the observed knockout effect was specific to the fear-responsive units identified by our selection procedure, or instead a generic consequence of removing neurons, perturbing particular layers, suppressing strongly activated units, or applying a selection procedure that would identify apparently important units even when no meaningful affective structure was present. Knockout of the identified fear-responsive units produced an observed effect of +0.129, exceeding the 95% empirical null interval for all four control families: same-$K$ random neurons (null mean +0.0029, interval [−0.0143, +0.0143]), layer-matched random neurons (+0.0036, [−0.0143, +0.0218]), activation-magnitude-matched non-emotion neurons (−0.0000, [−0.0729, +0.0429]) and shuffled-label selection (+0.0114, [−0.0143, +0.0504]). None of these alternative explanations reproduced the magnitude of the targeted knockout effect. Each 20-draw comparison returned $p = 0.048$, the resolution floor attainable with that many draws under Eq. (22). The pooled 80-draw null had mean +0.0045, interval [−0.0143, +0.0429] and $p = 0.012$, with the observed effect again lying well beyond the null range.

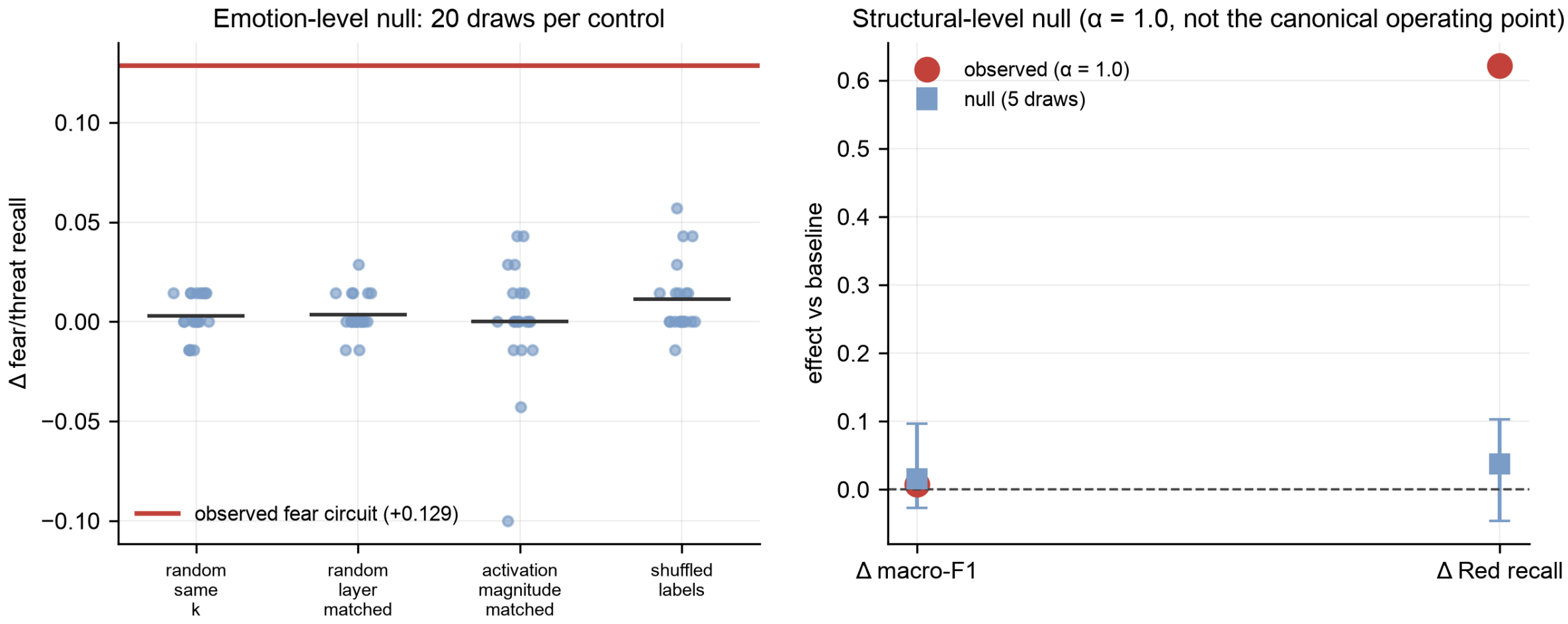

**Figure 19.** Specificity controls. Left: the source fear-circuit knockout exceeds the same-$K$, layer-matched, activation-matched and shuffled-label nulls, each estimated from 20 draws. Right: five structural random directions at $\alpha = 1.0$, shown as a boundary check; this is not an exact-strength null for the $\alpha = 0.5$ result.

The target-domain control evidence points in the same direction, but it is statistically weaker because there were too few random-control draws. The norm-matched direction at $\alpha = +0.5$ was null, as reported above. A separate five-draw family was run at $\alpha = 1.0$, where the Red-recall null mean was +0.038 with a maximum of +0.108 against an observed targeted effect of +0.622 ($p = 0.167$, purely because five draws cannot produce a smaller empirical value). Rejected conditions, happiness at $\alpha = -1.0$ and sadness at $\alpha = \pm 1.0$, in which steering caused 75% to 89% of Stage-1 outputs to hit the token limit, were treated as degeneration failures and excluded from mechanistic interpretation.

## 4.8 Sparse neurons and the distributed direction dissociate

The final question is which representational object carries the transfer. The sparse fear/threat neurons are demonstrably necessary for source emotion recognition (Section 4.3) but do not carry the building criterion shift: amplifying or ablating the 256-neuron set changed Red recall by only +0.027, against +0.486 for the distributed direction. The two source metrics are not a common effect size, since the sparse metric is a recall deficit and the distributed metric a prediction-tendency change, but the target dissociation is direct, because identical building cases respond strongly to one intervention and minimally to the other.

The train-fit overlap analysis converges on the same conclusion (Figure 20). Over the 56,832-neuron candidate-band universe, fear versus unsafe-decision overlap was 1 neuron against 1.15 expected (0.87-fold, $p = 0.686$), and fear versus damage-severity overlap was 0 against 1.15 expected ($p = 1.000$). The two structural sets, however, overlapped each other by 4 neurons, 3.47-fold above chance ($p = 0.029$), confirming that the method detects enrichment when enrichment exists. Sparse fear-selective units are therefore simply not part of the structural decision machinery.

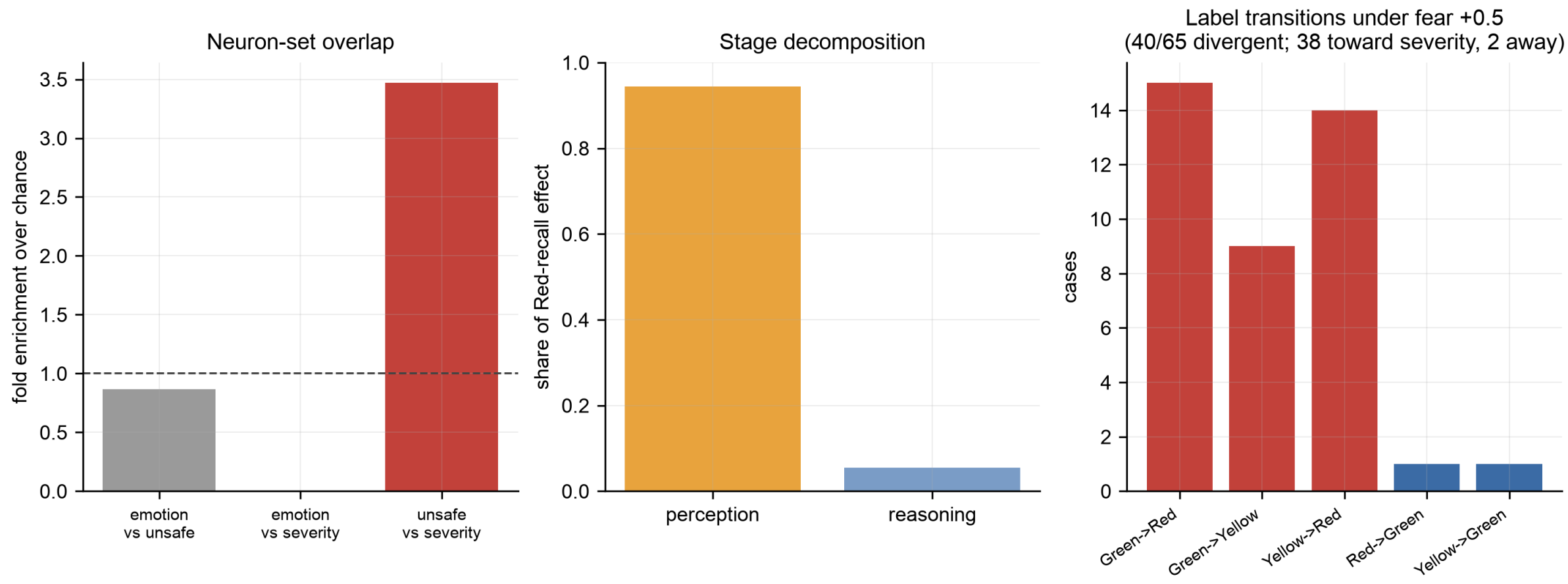

**Figure 20.** Circuit analysis. Sparse-set overlap is at chance for fear versus unsafe-decision and for fear versus damage-severity, but above chance between the two structural sets. The figure also summarises the stage decomposition and the strongly severity-increasing label transitions under positive fear/threat steering.

Overlap is associational and does not establish that a shared neuron transmits a causal effect. Taken with the stage result of Section 4.6 and the family dissociation above, the most plausible mechanism is a distributed representational shift in the residual stream during perception, which changes what the perceptual report says and then propagates through an unmodified reasoning stage. It is not a specific set of fear neurons wired into the safety decision.

# 5. Discussion

## 5.1 Criterion control as the engineering contribution

This study began from an observation that the disaster-informatics literature reports but does not treat as a design target: VLM safety assessors are reluctant to call a hazard. Zero-shot hazard recall below 0.33 for frontier multimodal models [7], 28% of real hazards undetected by a deployed agent [12], and the behaviour of our own reconstruction (Red recall 0.270 with Red precision 1.000) all describe the same failure. Our first contribution is to show that this is not a perception deficit. The pipeline separates unsafe from non-unsafe buildings at $d' = 1.521$ and places its boundary at $c = +1.354$; supplying ground-truth perception lifts $d'$ to 2.416 while leaving $c$ at +0.907. Better recognition, which is what most of this literature pursues [4,5,71], therefore would not have fixed the problem. The field reports accuracy at one operating point and leaves the operating point itself unexamined, which is precisely the gap that SDT was developed to close [57,72] and that construction-safety research has already exploited for human assessors [55,59].

Our second contribution is a method that moves that boundary from inside the model. A direction estimated entirely from decontaminated natural photographs, frozen before the test split was touched, shifted the Unsafe boundary by 1.515 units of standardised evidence and shifted it bidirectionally, with every weight and prompt unchanged. Four facts distinguish this from a perturbation that merely changes outputs: the direction was selected on independent data; a norm-matched random direction of identical magnitude was null ($p = 0.457$); a matched positive-valence direction was null ($p = 0.84$); and four families of matched sparse control were null at the source. Methodologically this extends contrastive activation steering [28–30]: prior works all identify and apply its control signal within the task being controlled, whereas we show that a representation validated on one task and frozen can transfer to an unrelated engineering decision. The affective axis was motivated by a human regularity (fear lowers the evidence threshold for declaring a situation dangerous [20,21,53]), and we found that the VLM's behaviour matched that prediction in sign, magnitude and specificity.

The SDT decomposition is not a caveat attached to that claim; it is the claim. Red recall rose by 0.486 and macro-F1 by 0.174, but $\Delta c = -1.515$ against $\Delta d' = -0.493$, a ratio of 3.1. The intervention made the model readier to act on evidence it already had and did not improve the evidence. One terminological clarification is needed because the relevant vocabularies point in opposite directions: in SDT terms the intervention makes the model more liberal, since less evidence is required before it responds Red, whereas in everyday engineering terms the same change makes it more cautious, since it closes more buildings. Both describe $\Delta c < 0$, and we use the SDT convention throughout.

Two practical consequences follow, and they cut in opposite directions. The first is a warning about evaluation. Figure 9(B) shows that sweeping a threshold over the baseline's own probabilities moves macro-F1 from 0.21 to 0.45 and Red recall from 0.27 to 1.00 with the evidence untouched. A single reported operating point therefore describes a policy, not a competence. Because the test split is Red-majority, our own macro-F1 gain is partly mechanical, and only one of 24 comparisons survives multiplicity correction. Any benchmark result in this literature that reports accuracy alone is open to the same confound, and reporting $d'$ and $c$ alongside conventional metrics costs nothing and prevents the error. The second is more encouraging. The obvious objection to an internal method is that one could simply threshold the output probability instead. Our data answer that directly: driven to the same 38 Red postings, external thresholding yields macro-F1 0.376 and $d' = 0.529$, whereas internal steering yields 0.556 and 1.027. Reaching a given criterion by re-weighting a poorly calibrated output distribution costs substantially more discrimination than reaching it by shifting the internal representation. For a deployment that must move its operating point and cannot retrain, that difference is the practical case for the method. Criterion placement nonetheless remains an engineering decision that must be made against an explicit loss function, the available capacity for human review, and the expected prevalence [73,74], and evaluated from the actual deployment baseline, since a pipeline already near its optimum would experience the same $\Delta c$ as harmful over-triage.

## 5.2 Mechanism: a distributed perceptual shift

Two results constrain the mechanism, and both are more specific than the behavioural effect alone would support. The first is the stage dissociation. Reasoning-only steering operated on Stage-1 text, parsed profiles, retrieved cases, retrieved rules and complete Stage-2 contexts confirmed byte-identical in all 65 cases, and recovered 5.6% of the effect, against 94.4% for perception-only steering. The vulnerable stage is the transformation of images into a structured engineering profile, not the conversion of fixed profile text and retrieved rules into a posting. This is actionable for modular pipelines of the kind now standard in disaster informatics [2,9]: it indicates where invariance should be monitored, and it shows that a deterministic rule layer downstream of a learned perception stage provides less protection than its determinism implies. The profile analysis also rules out the simplest story, since positive and negative steering both raised mean hazard severity while driving the posting in opposite directions, and the case-level correlation between severity change and decision change was only +0.308. What changes is the character of the perceptual report as a whole, consistent with evidence that visual information in VLMs is progressively re-expressed in vocabulary space across late layers and aggregated at the final prompt position [27].

The second is the sparse-versus-distributed dissociation, which is a contribution to the interpretability literature as well as to this application. The 256 selected neurons are reproducibly causal for source emotion recognition, selectively, against four null families including a shuffled-label re-run of the entire selection pipeline, but they transfer nothing. The overlap analysis explains why: they share essentially no membership with either the unsafe-decision or the damage-severity circuit, while those two overlap each other 3.5-fold above chance. This qualifies a growing body of work that localizes emotion to discrete units and validates them by ablation [32,33,35]: such neurons can be genuinely necessary for the task on which they were found and still be the wrong object for transfer. A residual-stream direction pools many weakly aligned features, including components well outside any selector's top ranks, and it is that pooled object which reaches the downstream computation, consistent with the linear, distributed encoding of affective variables reported for text models [31] and with the residual stream as the medium through which transformer components communicate [23]. The practical lesson is that demonstrating causal necessity on a source task licenses no inference about what will transfer, and that a neuron list is not a circuit.

## 5.3 Limitations and future work

The clearest limitation is scope. The experiments use one backbone, one affect corpus and one building benchmark, so they establish that a controllable criterion exists in this system before claim that it is general. We nonetheless expect the design to travel further than the result, because nothing in the localize–validate–freeze–transfer sequence is specific to Qwen2.5-VL, to FindingEmo or to earthquakes: it requires only a model whose residual stream can be read and written, a source task with contrastive labels that is decontaminated against the target domain, and a target task with a measurable decision. We would encourage replication across a second backbone and a second whole-scene affect corpus, across non-earthquake hazards such as hurricane, fire and flood damage where VLM pipelines are already deployed [9,11,13], and across the affective axes the present design could not test. The same scaffolding also applies to non-affective latent factors, for example, urgency, authority, optimism, fatigue, or demographic stereotype, any of which could in principle move an engineering decision criterion without changing evidence quality, and each of which could be audited the way we audit fear here.

Statistical power is the second limitation, given that the published test split contains 65 cases and is Red-heavy. Pairing and the event-cluster bootstrap reduce but cannot remove the resulting uncertainty, and only one of 24 macro-F1 comparisons survives multiplicity correction, so we rest the paper on the large bidirectional criterion change and its stage localization instead of on small accuracy differences, and we report the macro-F1 gain as partly mechanical. Control depth is uneven for the same reason: source specificity rests on four 20-draw null families and is strong, whereas the structural null used five draws and was evaluated at $\alpha = 1.0$ instead of $\alpha = 0.5$. Larger and event-disjoint benchmarks would help most, since earthquake-event overlap between train and test is a property of the published split that we retained unchanged.

Mechanistic completeness is the third limitation. Stage-resolved intervention, the sparse–distributed dissociation and set overlap jointly constrain the explanation, but overlap is associational and none of these traces a path. Activation patching is the natural next step, and the circuit analysis already identified the matched case pairs whose postings differ between baseline and steered runs. Patching emotion-sensitive MLP activations, candidate attention-head outputs and residual-stream states between baseline and steered runs would localise where the causal effect is actually transmitted, in the manner established for factual recall [25] and applied to emotion inference in text models [34]. Head-level attribution [39], a matched-layer comparison of the sparse and distributed families, and sparse-dictionary decomposition of the steering direction into interpretable features [26] would further test whether the distributed direction is a single feature or a bundle of them.

# 6. Conclusion

A post-earthquake safety pipeline built on Qwen2.5-VL-7B issues an Unsafe posting for only 27.0% of genuinely unsafe buildings while never issuing a false one. This is not a perception failure, as the pipeline separates unsafe from non-unsafe buildings at $d' = 1.521$, but a decision criterion sitting at $c = +1.354$, far from any position a hazard-averse deployment would choose. We introduced FearCaut-Qwen, which moves that criterion from inside the model at inference time using an affective direction localized on decontaminated natural scenes, causally validated by sparse knockout and distributed steering on held-out emotion data, and frozen before the structural test split was accessed.

Injecting the frozen fear/threat direction raised Red recall from 0.270 to 0.757 and subtracting it reduced Red recall to zero, with a norm-matched random direction and a matched happiness direction both null. The decomposition identifies the mechanism unambiguously: the criterion moved by $-1.515$ while discrimination was not improved ($\Delta d' = -0.493$), a ratio of 3.1. Reaching the same operating point by thresholding the model's own output probabilities instead cost far more discrimination ($d' = 0.529$ against 1.027), so the internal route is not merely a re-description of output calibration. Perception-only intervention reproduced 94.4% of the effect and reasoning-only intervention over byte-identical evidence reproduced 5.6%, placing the mechanism in the construction of the perceptual report rather than in rule-based reasoning. Sparse emotion neurons that were causally necessary for source recognition carried none of the transfer; a distributed residual-stream direction carried all of it.

For building engineering the message has two parts. A model's willingness to declare a building unsafe is a distinct system property from its ability to tell unsafe from safe; both must be measured, and the first can be set deliberately at inference without retraining. And because moving a badly placed threshold raises accuracy on an imbalanced benchmark while discrimination falls, any evaluation that reads accuracy alone will mistake a criterion shift for competence. Reporting $d'$ and $c$ alongside conventional metrics costs nothing and prevents that error.

# Data availability

SeisMLLM-1K is publicly available from its original release. FindingEmo distributes annotations and source URLs rather than image files; the annotation index, the image identifier lists for all three splits, the SHA-256 manifests and the decontamination decisions are reported with the study. All configuration files, resolved run metadata, prediction records, frozen circuit records with their content hashes, and the complete analysis code are available from the authors on request.